\documentclass[10pt,twocolumn,letterpaper]{article}

\usepackage{cvpr}
\usepackage{times}
\usepackage{epsfig}
\usepackage{graphicx}
\usepackage{amsmath}
\usepackage{amssymb}
\usepackage{comment}
\usepackage{xcolor}
\usepackage{soul}
\usepackage{array}
\usepackage{textcomp}
\usepackage{float}
\usepackage{pdfpages}
\newcolumntype{L}{>{\centering\arraybackslash}m{1.2cm}}
\newcolumntype{S}{>{\centering\arraybackslash}m{1.0cm}}
\newcolumntype{D}{>{\centering\arraybackslash}m{0.8cm}}
\usepackage{amssymb}
\usepackage{pifont}

\newcommand{\R}{\mathbb{R}}
\let \bs=\mathbf
\let \set=\mathcal

\def \saliency {\textup{\saliency}}

\def \init {\mathit{init}}
\def \path {\mathit{path}}

\def \pose {\textup{pose}}
\def \spec {\textup{spec}}

\newtheorem{proposition}{\textbf{Proposition}}

\let \set = \mathcal
\let \bs = \boldsymbol

\newcommand{\chensong}[1]{\textcolor{red}{chensong:{#1}}}
\newcommand{\qixing}[1]{\textcolor{blue}{Qixing:{#1}}}

\usepackage[pagebackref=true,breaklinks=true,letterpaper=true,colorlinks,bookmarks=false]{hyperref}

\cvprfinalcopy 

\def\cvprPaperID{1176} 

\ifcvprfinal\fi
\begin{document}

\title{HybridPose: 6D Object Pose Estimation under Hybrid Representations}

\author{Chen Song\thanks{Authors contributed equally} , Jiaru Song$^*$, Qixing Huang\\
The University of Texas at Austin\\
{\tt\small \hyperref[mailto:song@cs.utexas.edu]{song@cs.utexas.edu}, \hyperref[mailto:jiarus@cs.utexas.edu]{jiarus@cs.utexas.edu}, \hyperref[mailto:huangqx@cs.utexas.edu]{huangqx@cs.utexas.edu}}
}

\maketitle
\thispagestyle{empty}

\begin{abstract}
We introduce HybridPose, a novel 6D object pose estimation approach. HybridPose utilizes a hybrid intermediate representation to express different geometric information in the input image, including keypoints, edge vectors, and symmetry correspondences. Compared to a unitary representation, our hybrid representation allows pose regression to exploit more and diverse features when one type of predicted representation is inaccurate (e.g., because of occlusion). Different intermediate representations used by HybridPose can all be predicted by the same simple neural network, and outliers in predicted intermediate representations are filtered by a robust regression module. Compared to state-of-the-art pose estimation approaches, HybridPose is comparable in running time and accuracy. For example, on Occlusion Linemod~\cite{brachmann2014learning} dataset, our method achieves a prediction speed of 30 fps with a mean ADD(-S) accuracy of 47.5\%, representing a state-of-the-art performance\footnote{We are informed by readers that our previous experimental setup is inconsistent with baselines. This problem is fixed in the current version of the paper. Please refer to our GitHub issues for related discussions.}. The implementation of HybridPose is available at \href{https://github.com/chensong1995/HybridPose}{https://github.com/chensong1995/HybridPose}.
\end{abstract}

\section{Introduction}
\label{Section:Introduction}

Estimating the 6D pose of an object from an RGB image is a fundamental problem in 3D vision and has diverse applications in object recognition and robot-object interaction. Advances in deep learning have led to significant breakthroughs in this problem. While early works typically formulate pose estimation as end-to-end pose classification~\cite{TulsianiM15} or pose regression~\cite{kendall2015posenet,XiangSNF18}, recent pose estimation methods usually leverage keypoints as an intermediate representation~\cite{TekinSF18,pvnet}, and align predicted 2D keypoints with ground-truth 3D keypoints. In addition to ground-truth pose labels, these methods incorporate keypoints as an intermediate supervision, facilitating smooth model training. Keypoint-based methods are built upon two assumptions: (1) a machine learning model can accurately predict 2D keypoint locations; and (2) these predictions provide sufficient constraints to regress the underlying 6D pose. Both assumptions easily break in many real-world settings. Due to object occlusions and representational limitations of the prediction network, it is often impossible to accurately predict 2D keypoint coordinates from an RGB image alone. 


\begin{figure}
\centering
\includegraphics[width=1.0\columnwidth]{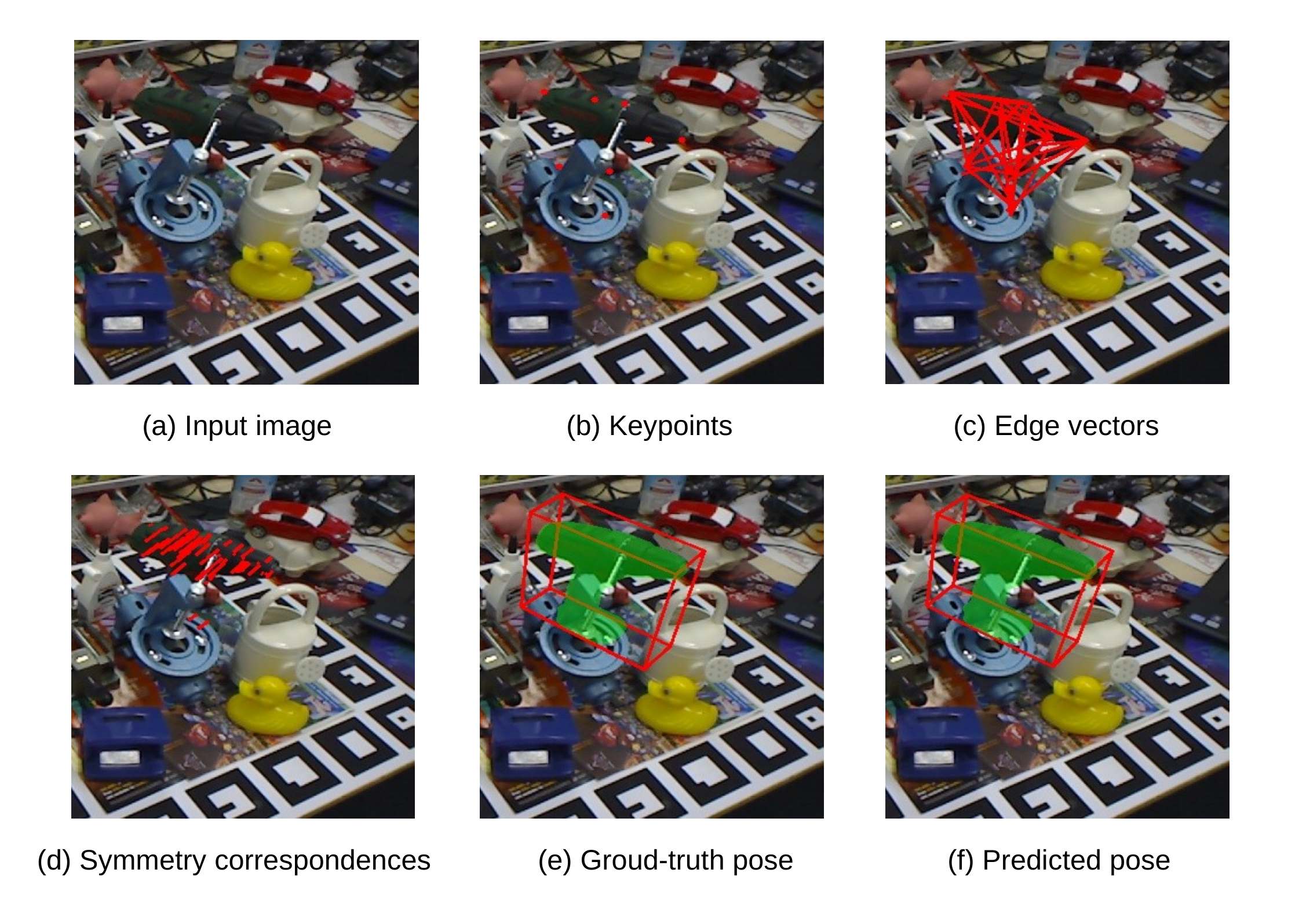}
\caption{\small{HybridPose predicts keypoints, edge vectors, and symmetry correspondences. In~(a), we show the input RGB image, in which the object of interest (driller) is partially occluded. In~(b), red markers denote predicted 2D keypoints. In~(c), edge vectors are defined by a fully-connected graph among all keypoints. In~(d), symmetry correspondences connect each 2D pixel on the object to its symmetric counterpart. For illustrative purposes, we only draw symmetry correspondences of 50 random samples from 5755 detected object pixels in this example. The predicted pose~(f) is obtained by jointly aligning all predictions with the 3D template, which involves solving a non-linear optimization problem.}}
\label{Figure:Teaser}
\vspace{-0.15in}
\end{figure}

In this paper, we introduce HybridPose, a novel 6D pose estimation approach that leverages multiple intermediate representations to express the geometric information in the input image. In addition to keypoints, HybridPose integrates a prediction network that outputs edge vectors between adjacent keypoints. As most objects possess a (partial) reflection symmetry, HybridPose also utilizes predicted dense pixel-wise correspondences that reflect the underlying symmetric relations between pixels. Compared to a unitary representation, this hybrid representation enjoys a multitude of advantages. First, HybridPose integrates more signals in the input image: edge vectors encode spacial relations among object parts, and symmetry correspondences incorporate interior details. Second, HybridPose offers more constraints than using keypoints alone for pose regression, enabling accurate pose prediction even if a significant fraction of predicted elements are outliers (e.g., because of occlusion). Finally, 
it can be shown that symmetry correspondences stabilize the rotation component of pose prediction, especially along the normal direction of the reflection plane (details are provided in the supp. material). 

Given the intermediate representation predicted by the first module, the second module of HybridPose performs pose regression. In particular, HybridPose employs trainable robust norms to prune outliers in predicted intermediate representation. We show how to combine pose initialization and pose refinement to maximize the quality of the resulting object pose. We also show how to train HybridPose effectively using a training set for the pose prediction module, and a validation set for the pose regression module.

We evaluate HybridPose on two popular benchmark datasets, Linemod~\cite{Hinterstoisser:2012:MBT:2481913.2481959} and Occlusion Linemod~\cite{brachmann2014learning}. In terms of accuracy (under the ADD(-S) metric), HybridPose leads to improvements from state-of-the-art methods that merely utilize keypoints. On Occlusion Linemod~\cite{brachmann2014learning}, HybridPose achieves an accuracy of 47.5\%, which beats DPOD~\cite{zakharov2019dpod}, the current state-of-the-art method on this benchmark dataset. 

Despite the gain in accuracy, our approach is efficient and runs at 30 frames per second on a commodity workstation. Compared to approaches which utilize sophisticated network architecture to predict one single intermediate representation  (such as Pix2Pose~\cite{park2019pix2pose}), HybridPose achieves better performance by using a relative simple network to predict hybrid representations.


\section{Related Works}
\label{Section:Related:Works}

\begin{figure*}
\centering
\includegraphics[width=1.0\textwidth]{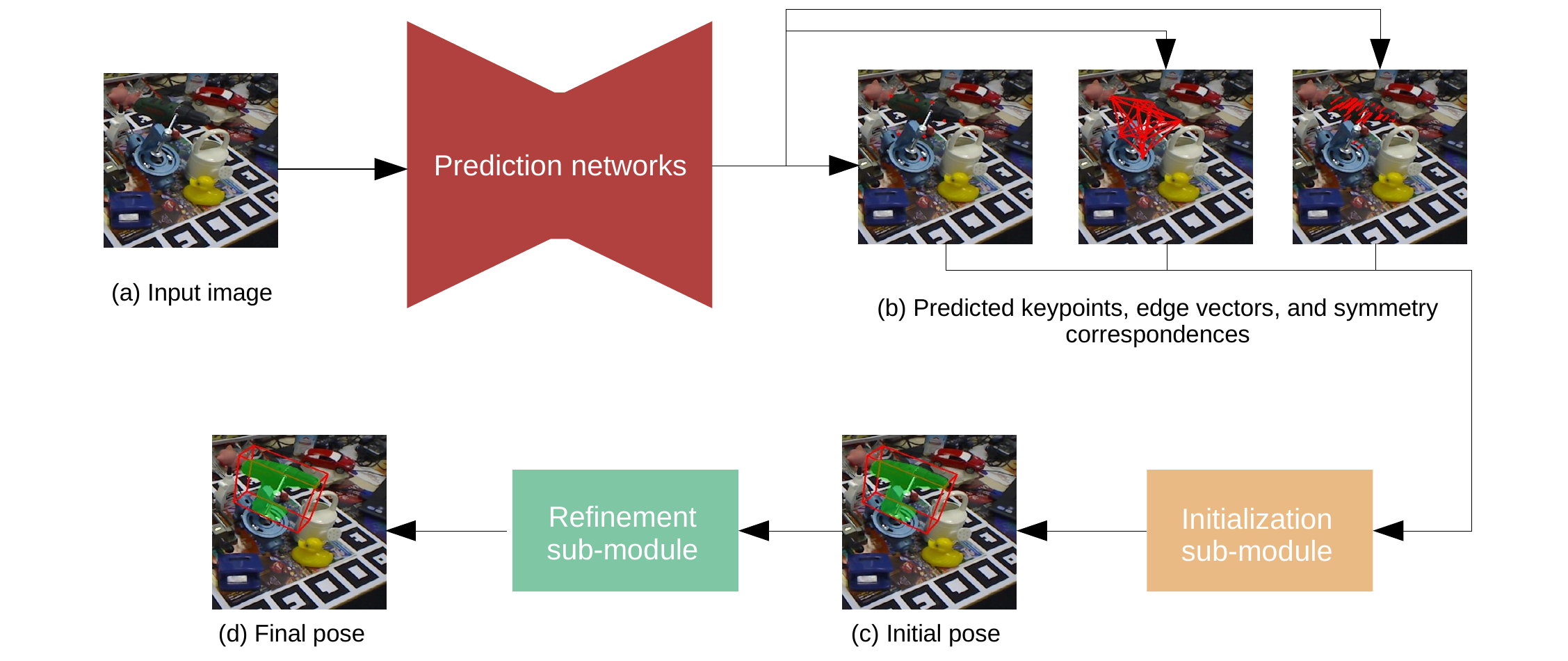}
\caption{\small{\textbf{Approach overview.} HybridPose consists of intermediate representation prediction networks and a pose regression module. The prediction networks take an image as input, and output predicted keypoints, edge vectors, and symmetry correspondences. The pose regression module consists of a initialization sub-module and a refinement sub-module. The initialization sub-module solves a linear system with predicted intermediate representations to obtain an initial pose. The refinement sub-module utilizes GM robust norm and optimizes (\ref{Eq:Pose}) to obtain the final pose prediction.} }
\label{Figure:Approach:Overview}
\vspace{-0.15in}
\end{figure*}

\noindent\textbf{Intermediate representation for pose.} To express the geometric information in an RGB image, a prevalent intermediate representation is keypoints, which achieves state-of-the-art performance~\cite{pvnet,2017icrapavlakos20176,rad2017bb8}. The corresponding pose estimation pipeline combines keypoint prediction and pose regression initialized by the PnP algorithm~\cite{lepetit2009epnp}. Keypoint predictions are usually generated by a neural network, and previous works use different types of tensor descriptors to express 2D keypoint coordinates. A common approach represents keypoints as peaks of heatmaps~\cite{NewellYD16,zhou2018starmap}, which becomes sub-optimal when keypoints are occluded, as the input image does not provide explicit visual cues for their locations. Alternative keypoint representations include vector-fields~\cite{pvnet} and patches~\cite{hu2019segmentation}. These representations allow better keypoint predictions under occlusion, and eventually lead to improvement in pose estimation accuracy. However, keypoints alone are a sparse representation of the object pose, whose potential in improving estimation accuracy is limited.

Besides keypoints, another common intermediate representation is the coordinate of every image pixel in the 3D physical world, which provides dense 2D-3D correspondences for pose alignment, and is robust under occlusion~\cite{brachmann2014learning,brachmann2016uncertainty,park2019pix2pose,li2019cdpn}. However, regressing dense object coordinates is much more costly than keypoint prediction. They are also less accurate than keypoints due to the lack of corresponding visual cues. In addition to keypoints and pixel-wise 2D-3D correspondences, depth is another alternative intermediate representation in visual odometry settings, which can be estimated together with pose in an unsupervised manner~\cite{zhou2017unsupervised}. In practice, the accuracy of depth estimation is limited by the representational power of neural networks. 

Unlike previous approaches, HybridPose combines multiple intermediate representations, and exhibits collaborative strength for pose estimation.

\noindent\textbf{Multi-modal input.} To address the challenges for pose estimation from a single RGB image, several works have considered inputs from multiple sensors. A popular approach is to leverage information from both RGB and depth images~\cite{zhou2017unsupervised,wang2019densefusion,XiangSNF18}. In the presence of depth information, pose regression can be reformulated as the 3D point alignment problem, which is then solved by the ICP algorithm~\cite{XiangSNF18}. Although HybridPose utilizes multiple intermediate representations, all intermediate representations are predicted from an RGB image alone. HybridPose handles situations in which depth information is absent.


\noindent\textbf{Edge features.}
Edges are known to capture important image features such as object contours~\cite{bertasius2015deepedge}, salient edges~\cite{liu2016learning}, and straight line segments~\cite{zhang2019ppgnet}. Unlike these low-level image features, HybridPose leverages semantic edge vectors defined between adjacent keypoints. This representation, which captures correlations between keypoints and reveals underlying structure of object, is concise and easy to predict. Such edge vectors offer more constraints than keypoints alone for pose regressions and have clear advantages under occlusion. Our approach is similar to~\cite{CaoSWS17}, which predicts directions between adjacent keypoints to link keypoints into a human skeleton. However, we predict both the direction and the magnitude of edge vectors, and use these vectors to estimate object poses.



\noindent\textbf{Symmetry detection from images.} 
Symmetry detection has received significant attention in computer vision. We refer readers to~\cite{LiuHKG10,Mitra:2013:SGE} for general surveys, and~\cite{AtadjanovL16,WangTZ15} for recent advances. Traditional applications of symmetry detection include face recognition~\cite{passalis2011using}, depth estimation~\cite{liu2016symmetrydepth}, and 3D reconstruction~\cite{hong2004symmetry,xue2011symmetric}. In the context of object pose estimation, people have studied symmetry from the perspective that it introduces ambiguities for pose estimation (c.f.~\cite{manhardt2019explaining,rad2017bb8, XiangSNF18}), since symmetric objects with different poses can have the same appearance in image. Several works~\cite{rad2017bb8,XiangSNF18,corona2018pose,manhardt2019explaining,park2019pix2pose} have explored how to address such ambiguities, e.g., by designing loss functions that are invariant under symmetric transformations. 

\noindent\textbf{Robust regression.} Pose estimation via intermediate representation is sensitive to outliers in predictions, which are introduced by occlusion and cluttered backgrounds~\cite{sundermeyer2018implicit,2017icrapavlakos20176,wang2019densefusion}. To mitigate pose error, several works assign different weights to different predicted elements in the 2D-3D alignment stage ~\cite{pvnet,2017icrapavlakos20176}. In contrast, our approach additionally leverages robust norms to automatically filter outliers in the predicted elements. 


Besides the reweighting strategy, some recent works propose to use deep learning-based refiners to boost the pose estimation performance~\cite{li2018deepim,manhardt2018deep,zakharov2019dpod}. \cite{zakharov2019dpod,li2018deepim} use point matching loss and achieve high accuracy. \cite{manhardt2018deep} predicts pose updates using contour information. Unlike these works, our approach considers the critical points and the loss surface of the robust objective function, and does not involve a fixed pre-determined iteration count used in recurrent network based approaches.


\section{Approach}

The input to HybridPose is an image $I$ containing an object in a known class, 
taken by a pinhole camera with known intrinsic parameters. Assuming that the class of objects has a canonical coordinate system $\Sigma$ (i.e. the 3D point cloud), HybridPose outputs the 6D camera pose $(R_I\in SO(3), \bs{t}_I \in \R^3)$ of the image object under $\Sigma$, where $R_I$ is the rotation and $\bs{t}_I$ is the translation component.

\subsection{Approach Overview}
\label{Subsec:Approach:Overview}

As illustrated in Figure~\ref{Figure:Approach:Overview}, HybridPose consists of a prediction module and a pose regression module. 

\noindent\textbf{Prediction module (Section~\ref{Subsec:Hybrid:Representation}).} HybridPose utilizes three prediction networks $f_{\theta}^{\set{K}}$, $f_{\phi}^{\set{E}}$, and $f_{\gamma}^{\set{S}}$ to estimate a set of keypoints $\set{K} = \{\bs{p}_k\}$, a set of edges between keypoints $\set{E} = \{\bs{v}_{e}\}$, and a set of symmetry correspondences between image pixels $\set{S} = \{(\bs{q}_{s,1}, \bs{q}_{s,2})\}$. $\set{K}$, $\set{E}$, and $\set{S}$ are all expressed in 2D. $\theta$, $\phi$, and $\gamma$ are trainable parameters.

The keypoint network $f_{\theta}^{\set{K}}$ employs an off-the-shelf prediction network~\cite{pvnet}. The other two prediction networks, $f_{\phi}^{\set{E}}$, and $f_{\gamma}^{\set{S}}$, are introduced to stabilize pose regression when keypoint predictions are inaccurate. Specifically, $f_{\phi}^{\set{E}}$ predicts edge vectors along a pre-defined graph of keypoints, which stabilizes pose regression when keypoints are cluttered in the input image. $f_{\gamma}^{\set{S}}$ predicts symmetry correspondences that reflect the underlying (partial) reflection symmetry. A key advantage of this symmetry representation is that the number of symmetry correspondences is large: every image pixel on the object has a symmetry correspondence. As a result, even with a large outlier ratio, symmetry correspondences still provide sufficient constraints for estimating the plane of reflection symmetry for regularizing the underlying pose. Moreover, symmetry correspondences incorporate more features within the interior of the underlying object than keypoints and edge vectors.

\noindent\textbf{Pose regression module (Section~\ref{Subsection:Pose:Regression}).} The second module of HybridPose optimizes the object pose $(R_I, \bs{t}_I)$ to fit the output of the three prediction networks. This module combines a trainable initialization sub-module and a trainable refinement sub-module. In particular, the initialization sub-module performs SVD to solve for an initial pose in the global affine pose space. The refinement sub-module utilizes robust norms to filter out outliers in the predicted elements for accurate object pose estimation.

\noindent\textbf{Training HybridPose (Section~\ref{Subsection:Network:Training}).} We train HybridPose by splitting the dataset into a training set and a validation set. We use the training set to learn the prediction module, and the validation set to learn the hyper-parameters of the pose regression module. We have tried training HybridPose end-to-end using one training set. However, the difference between the prediction distributions on the training set and testing set leads to sub-optimal generalization performance. 



\subsection{Hybrid Representation}
\label{Subsec:Hybrid:Representation}

This section describes three intermediate representations used in HybridPose.

\noindent\textbf{Keypoints.} The first intermediate representation consists of keypoints, which have been widely used for pose estimation. Given the input image $I$, we train a neural network $f_{\theta}^{\set{K}}(I)\in \R^{2\times |\set{K}|}$ to predict 2D coordinates of a pre-defined set of $|\set{K}|$ keypoints. 
In our experiments, HybridPose incorporates an off-the-shelf architecture called PVNet~\cite{pvnet}, which is the state-of-the-art keypoint-based pose estimator that employs a voting scheme to predict both visible and invisible keypoints. 

Besides outliers in predicted keypoints, another limitation of keypoint-based techniques is that when the difference (direction and distance) between adjacent keypoints characterizes important information of the object pose, inexact keypoint predictions incur large pose error. 

\noindent\textbf{Edges.} The second intermediate representation, which consists of edge vectors along a pre-defined graph, explicitly models the displacement between every pair of keypoints. As illustrated in Figure~\ref{Figure:Approach:Overview}, HybridPose utilizes a simple network $f_{\phi}^{\set{E}}(I)\in \R^{2\times |\set{E}|}$ to predict edge vectors in the 2D image plane, where $|\set{E}|$ denotes the number of edges in the pre-defined graph. In our experiments, $\set{E}$ is a fully-connected graph, i.e., $|\set{E}| = \frac{|\set{K}| \cdot (|\set{K}|-1)}{2}$.



\noindent\textbf{Symmetry correspondences.} The third intermediate representation consists of predicted pixel-wise symmetry correspondences that reflect the underlying reflection symmetry. In our experiments, HybridPose extends the network architecture of FlowNet 2.0~\cite{IlgMSKDB17} that combines a dense pixel-wise flow and the semantic mask predicted by PVNet. The resulting symmetry correspondences are given by predicted pixel-wise flow within the mask region. Compared to the first two representations, the number of symmetry correspondences is significantly larger, which provides rich constraints even for occluded objects. However, symmetry correspondences only constrain two degrees of freedom in the rotation component of the object pose (c.f.~\cite{Ma:2003:IVI}). It is necessary to combine symmetry correspondences with other intermediate representations.

A 3D model may possess multiple reflection symmetry planes. For these models, we train HybridPose to predict symmetry correspondences with respect to the most salient reflection symmetry plane, i.e., one with the largest number of symmetry correspondences on the original 3D model.

\noindent\textbf{Summary of network design.} In our experiments, $f_{\theta}^{\set{K}}(I)$, $f_{\phi}^{\set{E}}(I)$, and $f_{\gamma}^{\set{S}}$ are all based on ResNet~\cite{He2015}, and the implementation details are discussed in Section~\ref{Section:Experimental:Setup}. Trainable parameters are shared across all except the last convolutional layer. Therefore, the overhead of introducing the edge prediction network $f_{\phi}^{\set{E}}(I)$ and the symmetry prediction network $f_{\gamma}^{\set{S}}$ is insignificant.




\subsection{Pose Regression}
\label{Subsection:Pose:Regression}
The second module of HybridPose takes predicted intermediate representations $\{\set{K},\set{E},\set{S}\}$ as input and outputs a 6D object pose $(R_I\in SO(3), \bs{t}_I\in \R^3)$ for the input image $I$. Similar to state-of-the-art pose regression approaches~\cite{Persson_2018_ECCV}, HybridPose combines an initialization sub-module and a refinement sub-module. Both sub-modules leverage all predicted elements. The refinement sub-module additionally leverages a robust function to model outliers in the predicted elements. 



In the following, we denote 3D keypoint coordinates in the canonical coordinate system as $\overline{\bs{p}}_k, 1\leq k \leq |\set{K}|$. To make notations uncluttered, we denote output of the first module, i.e., predicted keypoints, edge vectors, and symmetry correspondences as $\bs{p}_k\in \R^2, 1\leq k \leq |\set{K}|$, $\bs{v}_{e}\in \R^2, 1\leq e \leq |\set{E}|$, and  $(\bs{q}_{s,1}\in \R^2,\bs{q}_{s,2}\in \R^2), 1\leq s \leq |\set{S}|$, respectively. Our formulation also uses the homogeneous coordinates $\hat{\bs{p}}_k\in \R^3$, $\hat{\bs{v}_{e}}\in \R^3$,$\hat{\bs{q}}_{s,1}\in \R^3$ and $\hat{\bs{q}}_{s,2}\in \R^3$ of $\bs{p}_k$, $\bs{v}_{e}$, $\bs{q}_{s,1}$ and $\bs{q}_{s,2}$ respectively. The homogeneous coordinates are normalized by the camera intrinsic matrix.

\noindent\textbf{Initialization sub-module.} This sub-module leverages constraints between $(R_I, \bs{t}_I)$ and predicted elements and solves $(R_i,\bs{t}_I)$ in the affine space, which are then projected to $SE(3)$ in an alternating optimization manner. To this end, we introduce the following difference vectors for each type of predicted elements:
\begin{align}
\overline{\bs{r}}_{R, \bs{t}}^{\set{K}}(\bs{p}_k) & := \hat{\bs{p}}_k\times (R\overline{\bs{p}}_k+\bs{t}), \label{Eq:1} \\
\overline{\bs{r}}_{R,\bs{t}}^{\set{E}}(\bs{v}_e, \bs{p}_{e_s}) & := \hat{\bs{v}}_e\times (R\overline{\bs{p}}_{e_t}+\bs{t}) + \hat{\bs{p}}_{e_s}\times (R\overline{\bs{v}}_e) \label{Eq:2} \\
r^{\set{S}}_{R,\bs{t}}(\bs{q}_{s,1},\bs{q}_{s,2}) & := (\hat{\bs{q}}_{s,1}\times \hat{\bs{q}}_{s,2})^T \bs{R}\overline{\bs{n}}_{r}.\label{Eq:3}
\end{align}
where $e_s$ and $e_t$ are end vertices of edge $e$, $\overline{\bs{v}}_e = \overline{\bs{p}}_{e_t} - \overline{\bs{p}}_{e_s}\in \R^3$, and $\overline{\bs{n}}_r\in \R^3$ is the normal of the reflection symmetry plane in the canonical system. 

HybridPose modifies the framework of EPnP~\cite{lepetit2009epnp} to generate the initial poses. By combining these three constraints from predicted elements, we generate a linear system of the form $A\bs{x} = \bs{0}$, where $A$ is matrix and its dimension is $(3|\set{K}| + 3|\set{E}| + |\set{S}|)\times 12$. $\bs{x} = [\bs{r}_1^{\mathrm{T}},\bs{r}_2^{\mathrm{T}},\bs{r}_3^{\mathrm{T}},\bs{t}^{\mathrm{T}}]_{12\times1}^{\mathrm{T}}$ is a vector that contains rotation and translation parameters in affine space. To model the relative importance among keypoints, edge vectors, and symmetry correspondences, we rescale (\ref{Eq:2}) and (\ref{Eq:3}) by hyper-parameters $\alpha_{E}$ and $\alpha_{S}$, respectively, to generate $A$. 

Following EPnP~\cite{lepetit2009epnp}, we compute $\bs{x}$ as 
\begin{equation}
    \bs{x} = \sum_{i = 1}^N\gamma_i\bs{v}_i 
\end{equation}
where $\bs{v}_i$ is the $i_{\mathrm{th}}$ smallest right singular vector of $A$. Ideally, when predicted elements are noise-free, $N = 1$ with $\bs{x} = \bs{v}_1$ is an optimal solution. However, this strategy performs poorly given noisy predictions. Same as EPnP~\cite{lepetit2009epnp}, we choose $N = 4$. To compute the optimal $\bs{x}$, we optimize latent variables $\gamma_i$ and the rotation matrix $R$ in an alternating optimization procedure with following objective function:
\begin{equation}
\underset{R\in \R^{3\times 3},\gamma_i}{\min} \| \sum_{i = 1}^4\gamma_i R_i - R\|_{\set{F}}^2
\label{Eq:Init:objective}
\end{equation}
where $R_i \in \R^{3\times 3}$ is reshaped from the first $9$ elements of $\bs{v}_i$. After obtaining optimal $\gamma_i$, we project the resulting affine transformation $\sum_{i = 1}^4\gamma_i R_i$ into a rigid transformation. Due to space constraint, we defer details to the supp. material.

\noindent\textbf{Refinement sub-module.} Although (\ref{Eq:Init:objective}) combines hybrid intermediate representations and admits good initialization, it does not directly model outliers in predicted elements. Another limitation comes from (\ref{Eq:1}) and (\ref{Eq:2}), which do not minimize the projection errors (i.e., with respect to keypoints and edges), which are known to be effective in keypoint-based pose estimation (c.f.~\cite{Persson_2018_ECCV}). 

Benefited from having an initial object pose $(R^{\init},\bs{t}^{\init})$, the refinement sub-module performs local optimization to refine the object pose. We introduce two difference vectors that involve projection errors: $\forall k, e, s,$
\begin{align}
\bs{r}^{\set{K}}_{R,\bs{t}}(\bs{p}_k) &:= \set{P}_{R,\bs{t}}(\overline{\bs{p}}_k) - \bs{p}_k, \label{Eq:4} \\   
\bs{r}^{\set{E}}_{R,\bs{t}}(\bs{v}_e) &:=\set{P}_{R,\bs{t}}(\overline{\bs{p}}_{e_t})-\set{P}_{R,\bs{t}}(\overline{\bs{p}}_{e_s}) - \bs{v}_e, \label{Eq:5} 
\end{align}
where $\set{P}_{R,\bs{t}}:\R^3\rightarrow \R^2$ is the projection operator induced from the current pose $(R,\bs{t})$. 

To prune outliers in the predicted elements, we consider a generalized German-Mcclure (or GM) robust function
\begin{align}
\rho(x, \beta):= \beta_1^2/(\beta_2^2 + x^{2}). 
\end{align}
With this setup, HybridPose solves the following non-linear optimization problem for pose refinement:
\begin{align}
\min\limits_{R,\bs{t}} & \ \sum\limits_{k=1}^{|\set{K}|} \rho(\|\bs{r}^{\set{K}}_{R,\bs{t}}(\bs{p}_k)\|,\beta_{\set{K}}) \|\bs{r}^{\set{K}}_{R,\bs{t}}(\bs{p}_k)\|_{\Sigma_k}^2 \nonumber \\
& + \frac{|\set{K}|}{|\set{E}|}\sum\limits_{e=1}^{|\set{E}|}
\rho(\|\bs{r}^{\set{E}}_{R,\bs{t}}(\bs{v}_e)\|,\beta_{\set{E}}) \|\bs{r}^{\set{E}}_{R,\bs{t}}(\bs{v}_e)\|_{\Sigma_e}^2
\nonumber \\
& \ + \frac{|\set{K}|}{|\set{S}|}\sum\limits_{s=1}^{|\set{S}|}\rho(r^{\set{S}}_{R,\bs{t}}(\bs{q}_{s,1},\bs{q}_{s,2}), \beta_{\set{S}})r^{\set{S}}_{R,\bs{t}}(\bs{q}_{s,1},\bs{q}_{s,2})^2
\label{Eq:Pose}
\end{align}
where $\beta_{\set{K}}$, $\beta_{\set{E}}$, and $\beta_{\set{S}}$ are separate hyper-parameters for keypoints, edges, and symmetry correspondences. $\Sigma_k$ and $\Sigma_e$ denote the covariance information attached to the keypoint and edge predictions. $\|\bs{x}\|_{A} = (\bs{x}^T A \bs{x})^{\frac{1}{2}}$. When covariances of predictions are unavailable, we simply set $\Sigma_k = \Sigma_e = I_2$. The above optimization problem is solved by Gauss-Newton method starting from $R^{\init}$ and $\bs{t}^{\init}$.


In the supp. material, we provide a stability analysis of (\ref{Eq:Pose}), and show how the optimal solution of (\ref{Eq:Pose}) changes with respect to noise in predicted representations. We also show collaborative strength among all three intermediate representations. While keypoints significantly contribute to the accuracy of $\bs{t}$, edge vectors and symmetry correspondences can stablize the regression of $R$. 
\subsection{HybridPose Training}
\label{Subsection:Network:Training}

This section describes how to train the prediction networks and hyper-parameters of HybridPose using a labeled dataset $\set{T} = \{I, (\set{K}_I^{gt}, \set{E}_I^{gt}, \set{S}_I^{gt}, (R_I^{gt},\bs{t}_I^{gt}))\}$. With $I$, $\set{K}_I^{gt}$, $\set{E}_I^{gt}$, $\set{S}_I^{gt}$, and $(R_I^{gt},\bs{t}_I^{gt})$, we denote the  RGB image, labeled keypoints, edges, symmetry correspondences, and ground-truth object pose, respectively. A popular strategy is to train the entire model end-to-end, e.g., using recurrent networks to model the optimization procedure and introducing loss terms on the object pose output as well as the intermediate representations. However, we found this strategy sub-optimal. The distribution of predicted elements on the training set differs from that on the testing set. Even by carefully tuning the trade-off between supervisions on predicted elements and the final object pose, the pose regression model, which fits the training data, generalizes poorly on the testing data. 

Our approach randomly divides the labeled set $\set{T} = \set{T}_{train} \cup \set{T}_{val}$ into a training set and a validation set. $\set{T}_{train}$ is used to train the prediction networks, and $\set{T}_{val}$ trains the hyper-parameters of the pose regression model. Implementation and training details of the prediction networks are presented in Section~\ref{Section:Experimental:Setup}. In the following, we focus on training the hyper-parameters using $\set{T}_{val}$.

\noindent\textbf{Initialization sub-module.} Let $R_I^{\init}$ and $\bs{t}_I^{\init}$ be the output of the initialization sub-module. We obtain the optimal hyper-parameters $\alpha_{E}$ and $\alpha_{S}$ by solving the following optimization problem:
\begin{equation}
\min\limits_{\alpha_{E}, \alpha_{S}} \ \sum\limits_{I\in \set{T}_{\pose}} \ \big(\|R_I^{\init}-R_I^{gt}\|_{\set{F}}^2 + \|\bs{t}_I^{\init}-\bs{t}_I^{gt}\|^2\big).
\label{Eq:Pose:Training}
\end{equation}
Since the number of hyper-parameters is rather small, and the pose initialization step does not admit an explicit expression, we use the finite-difference method to compute numerical gradient, i.e., by fitting the gradient to samples of the hyper-parameters around the current solution. We then apply back-track line search for optimization.

\begin{figure*}
\includegraphics[width=1.0\textwidth]{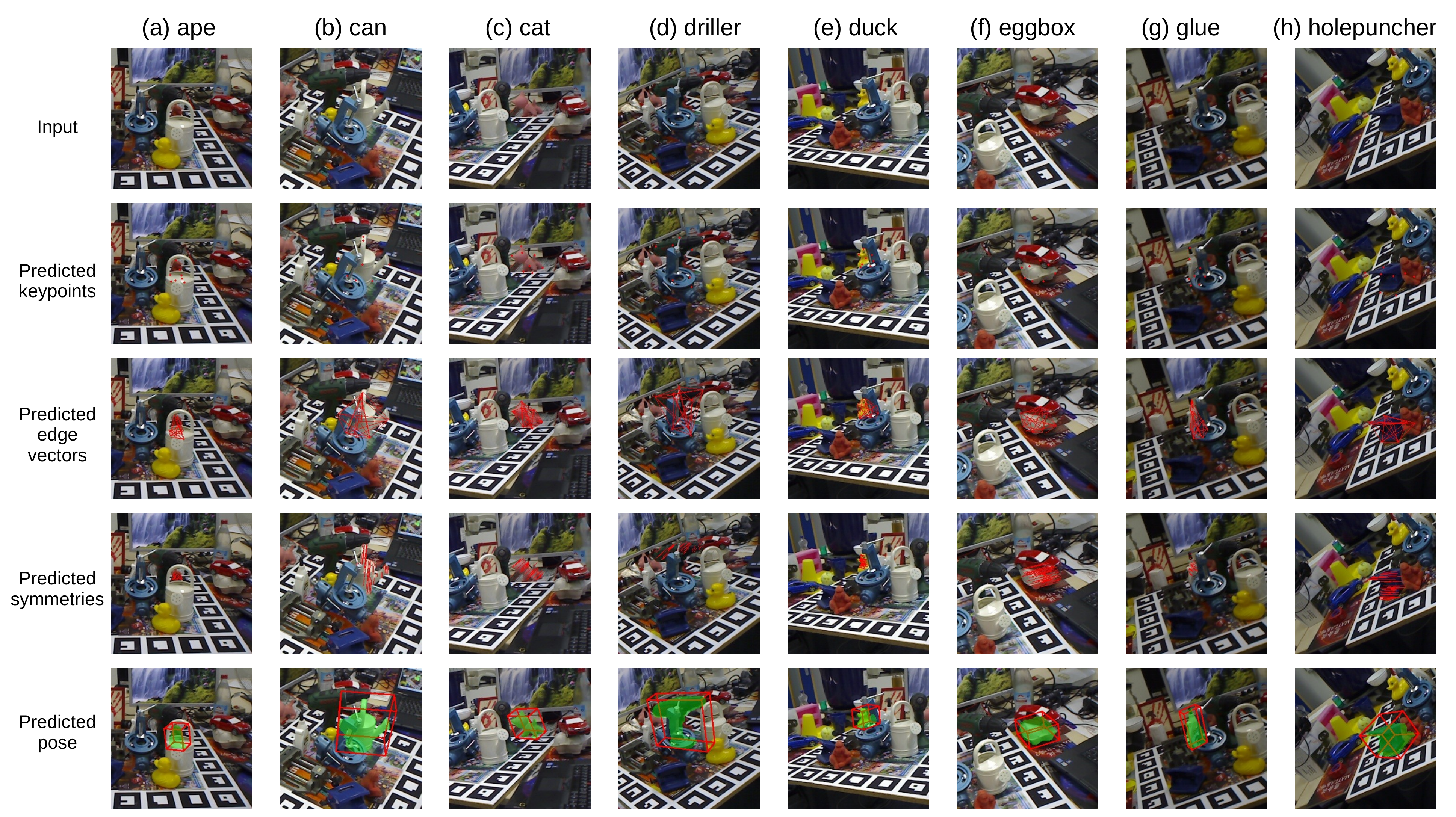}
\caption{\small{\textbf{Pose regression results.} HybridPose is able to accurately predict 6D poses from RGB images. HybridPose handles situations where the object has no occlusion (c), light occlusion (b, e, f, h), and severe occlusion (a, d, g). For illustrative purposes, we only draw 50 randomly selected symmetry correspondences in each example. }}
\label{Figure:Pose:Regression:Result}
\vspace{-0.05in}
\end{figure*}

\noindent\textbf{Refinement sub-module.} Let $\beta = \{\beta_{\set{K}},\beta_{\set{E}},\beta_{\set{S}}\}$ be the hyper-parameters of this sub-module. For each instance $(I, (\set{K}_I^{gt}, \set{E}_I^{gt}, \set{S}_I^{gt}, (R_I^{gt},\bs{t}_I^{gt})))\in \set{T}_{val}$, denote the objective function in (\ref{Eq:Pose})  as $f_I(\bs{c},\beta)$, where $\bs{c} = (\bs{c}^T,\overline{\bs{c}}^T)^T \in \R^6$ is a local parameterization of $R_I$ and $\bs{t}_I$, i.e., 
$
R_I = \exp(\bs{c}\times ) R_I^{gt}, \bs{t}_I = \bs{t}_I^{gt}+\overline{\bs{c}}
$. $\bs{c}$ encodes the different the current estimated pose and the ground-truth pose in $SE(3)$.

The refinement module solves an unconstrained optimization problem, whose optimal solution is dictated by its critical points and the loss surface around the critical points. We consider two simple objectives. The first objective forces $\frac{\partial f_I}{\partial \bs{c}}(\bs{0},\beta)\approx 0$, or in other words, the ground-truth is approximately a critical point. The second objective minimizes the condition number $\kappa(\frac{\partial^2 f_I}{\partial^2 \bs{c}}(\bs{0},\beta)) = \lambda_{\max}\big(\frac{\partial^2 f_I}{\partial^2 \bs{c}}(\bs{0},\beta)\big)/\lambda_{\min}\big(\frac{\partial^2 f_I}{\partial^2 \bs{c}}(\bs{0},\beta)\big)$. This objective regularizes the loss surface around each optimal solution, promoting a large converge radius for $f_I(\bs{c},\beta)$.  
With this setup, we formulate the following objective function to optimize $\beta$:
\begin{equation}
\min\limits_{\beta} \ \sum\limits_{I \in \set{T}_{val}} \|\frac{\partial f_I}{\partial \bs{c}}(\bs{0},\beta)\|^2 + \gamma \kappa\big(\frac{\partial^2 f_I}{\partial^2 \bs{c}}(\bs{0},\beta)\big) \label{Eq:Beta:Opt}
\end{equation}
where $\gamma$ is a constant hyperparemeter. The same strategy used in (\ref{Eq:Pose:Training}) is then applied to optimize (\ref{Eq:Beta:Opt}).

\section{Experimental Evaluation}

This section presents an experimental evaluation of the proposed approach. Section~\ref{Section:Experimental:Setup} describes the experimental setup. Section~\ref{Section:Analysis:Results} quantitatively and qualitatively compares HybridPose with other 6D pose estimation methods. Section~\ref{Section:Ablation:Study} presents an ablation study to investigate the effectiveness of symmetry correspondences, edge vectors, and the refinement sub-module.

\subsection{Experimental Setup}
\label{Section:Experimental:Setup}

\noindent\textbf{Datasets.} We consider two popular benchmark datasets that are widely used in the 6D pose estimation problem, Linemod~\cite{Hinterstoisser:2012:MBT:2481913.2481959} and Occlusion Linemod~\cite{brachmann2014learning}. In comparsion to Linemod, Occlusion Linemod contains more examples where the objects are under occlusion. Our keypoint annotation strategy follows that of~\cite{pvnet}, i.e., we choose $|\set{K}|=8$ keypoints via the farthest point sampling algorithm. Edge vectors are defined as vectors connecting each pair of keypoints. In total, each object has $|\set{E}| = \frac{|\set{K}| \cdot (|\set{K}|-1)}{2}=28$ edges. We further use the algorithm proposed in~\cite{ecins2018seeing} to annotate Linemod and Occlusion Linemod with reflection symmetry labels.

Following the convention described in~\cite{brachmann2016uncertainty}, we select 15\% of Linemod examples as the training data, and the rest 85\% as well as all of Occlusion Linemod examples for testing. To avoid overfitting, we use the same synthetic data generation scheme introduced in PVNet~\cite{pvnet}. \footnote{A previous version of this paper uses a different dataset split, which is inconsistent with baseline approaches. This problem has been fixed now. Please refer to our GitHub issues page for related discussions.}

\noindent\textbf{Implementation details.} We use ResNet~\cite{He2015} with pre-trained weights on ImageNet~\cite{deng2009imagenet} to build the prediction networks $f_{\theta}^{\set{K}}$, $f_{\phi}^{\set{E}}$, and $f_{\gamma}^{\set{S}}$. The prediction networks take an RGB image $I$ of size $(3, H, W)$ as input, and output a tensor of size $(C, H, W)$, where $(H, W)$ is the image resolution, and $C = 1 + 2|\set{K}| + 2|\set{E}| + 2$ is the number of channels in the output tensor.

The first channel in the output tensor is a binary segmentation mask $M$. If $M(x, y)=1$, then $(x, y)$ corresponds to a pixel on the object of interest in the input image $I$. The segmentation mask is trained using the cross-entropy loss.

The $2|\set{K}|$ channels afterwards in the output tensor give $x$ and $y$ components of all $|\set{K}|$ keypoints. A voting-based keypoint localization scheme~\cite{pvnet} is applied to extract the coordinates of 2D keypoints from this $2|\set{K}|$-channel tensor and the segmentation mask $M$. 


The next $2|\set{E}|$ channels in the output tensor give the $x$ and $y$ components of all $|\set{E}|$ edges, which we denote as $Edge$. Let $i$ ($0 \leq i < |\set{E}|$) be the index of an edge. Then
\small
\begin{equation*}
Edge_i = \{( Edge(2i, x, y), Edge(2i+1, x, y) )| M(x, y)= 1\}
\end{equation*}
\normalsize
is a set of 2-tuples containing pixel-wise predictions of the $i^{th}$ edge vector in $Edge$. The mean of $Edge_i$ is extracted as the predicted edge. 


The final 2 channels in the output tensor define the $x$ and $y$ components of symmetry correspondences. We denote this 2-channel ``map'' of symmetry correspondences as $Sym$. Let $(x, y)$ be a pixel on the object of interest in the input image, i.e. $M(x, y) = 1$. Assuming $\Delta x = Sym(0, x, y)$ and $\Delta y = Sym(1, x, y)$, we consider $(x, y)$ and $(x + \Delta x, y + \Delta y)$ to be symmetric with respect to the reflection symmetry plane. 

We train all three intermediate representations using the smooth $\ell_1$ loss  described in~\cite{Girshick_2015}.
Network training employs the Adam~\cite{kingma2014adam} optimizer for 200 epochs. The learning rate is set to $0.001$. Training weights of the segmentation mask, keypoints, edge vectors, and symmetry correspondences are 1.0, 10.0, 0.1, and 0.1, respectively.

The architecture described above achieves good performance in terms of detection accuracy. Nevertheless, it should be emphasized that the framework of HybridPose can incorporate future improvements in keypoint, edge vector, and symmetry correspondence detection techniques. 
Besides, Hybridpose can be extended to handling multiple objects within an image. One approach is to predict instance-level rather than semantic-level segmentation masks by methods such as Mask R-CNN~\cite{He_2017}. Intermediate representations are then extracted from each instance, and fed to the pose regression module in \ref{Subsection:Pose:Regression}.


\noindent\textbf{Evaluation protocols.} We use two metrics to evaluate the performance of HybridPose:

\noindent1. ADD(-S)~\cite{Hinterstoisser:2012:MBT:2481913.2481959, XiangSNF18} first calculates the distance between two point sets transformed by predicted pose and ground-truth pose respectively, and then extracts the mean distance. When the object possesses symmetric pose ambiguity, the mean distance is computed from the closest points between two transformed sets. ADD(-S) accuracy is defined as the percentage of examples whose calculated mean distance is less than 10\% of the model diameter.

\noindent2. In the ablation study, we compute and report the the angular rotation error $\|\frac{\log(R_{gt}^T R_I)}{2}\|$ and the relative translation error $\frac{\|\bs{t}_I-\bs{t}_{gt}\|}{d}$ between the predicted pose $(R_I,\bs{t}_I)$ and the ground-truth pose $(R_{gt},\bs{t}_{gt})$, where $d$ is object diameter.

\subsection{Analysis of Results}
\label{Section:Analysis:Results}

\setlength\tabcolsep{2pt}
\begin{table}
\centering
\footnotesize
\begin{tabular}{c | c c c c c c | c}
\hline
object           & Tekin & BB8  & Pix2Pose & PVNet          & CDPN          & DPOD           & Ours \\
\hline
ape              & 21.6  & 40.4 & 58.1     & 43.6           & 64.4          & \textbf{87.7}  & 63.1\\ 
benchvise        & 81.8  & 91.8 & 91.0     & 99.9           & 97.8          & 98.5           & \textbf{99.9}\\
cam              & 36.6  & 55.7 & 60.9     & 86.9           & 91.7          & \textbf{96.1}  & 90.4\\
can              & 68.8  & 64.1 & 84.4     & 95.5           & 95.9          & \textbf{99.7}  & 98.5\\
cat              & 41.8  & 62.6 & 65.0     & 79.3           & 83.8          & \textbf{94.7}  & 89.4\\
driller          & 63.5  & 74.4 & 76.3     & 96.4           & 96.2          & \textbf{98.8}  & 98.5\\
duck             & 27.2  & 44.3 & 43.8     & 52.6           & 66.8          & \textbf{86.3}  & 65.0\\
eggbox$^\dagger$ & 69.6  & 57.8 & 96.8     & 99.2           & 99.7          & 99.9           & \textbf{100.0}\\
glue$^\dagger$   & 80.0  & 41.2 & 79.4     & 95.7           & \textbf{99.6} & 96.8           & 98.8\\
holepuncher      & 42.6  & 67.2 & 74.8     & 81.9           & 85.8          & 86.9           & \textbf{89.7}\\
iron             & 75.0  & 84.7 & 83.4     & 98.9           & 97.9          & \textbf{100.0} & \textbf{100.0}\\
lamp             & 71.1  & 76.5 & 82.0     & 99.3           & 97.9          & 96.8           & \textbf{99.5}\\
phone            & 47.7  & 54.0 & 45.0     & 92.4           & 90.8          & 94.7           & \textbf{94.9} \\
\hline
average          & 56.0  & 62.7 & 72.4     & 86.3           & 89.9          & \textbf{95.2}  & 91.3 \\
\hline
\end{tabular}
\caption{\small{\textbf{Quantitative evaluation: ADD(-S) accuracy on Linemod.} Baseline approaches: Tekin et al.~\cite{TekinSF18}, BB8~\cite{rad2017bb8}, Pix2Pose~\cite{park2019pix2pose}, PVNet~\protect\cite{pvnet}, CDPN~\cite{li2019cdpn}, and DPOD~\cite{zakharov2019dpod}. Objects annotated with ($\dagger$) possess symmetric pose ambiguity. }}
\label{Figure:Qualitative:Lindmod}
\vspace{-0.12in}
\end{table}

\setlength\tabcolsep{2pt}
\begin{table}
\centering
\footnotesize
\begin{tabular}{c| c c c c c c | c }
\hline
object           & PoseCNN & Oberweger & Hu   & Pix2Pose & PVNet          & DPOD  & Ours \\
\hline
ape              & 9.6     & 12.1      & 17.6 & 22.0     & 15.8           & -     & \textbf{20.9}\\ 
can              & 45.2    & 39.9      & 53.9 & 44.7     & 63.3           & -     & \textbf{75.3}\\ 
cat              & 0.93    & 8.2       & 3.3  & 22.7     & 16.7           & -     & \textbf{24.9}\\ 
driller          & 41.4    & 45.2      & 62.4 & 44.7     & 65.7           & -     & \textbf{70.2}\\ 
duck             & 19.6    & 17.2      & 19.2 & 15.0     & 25.2           & -     & \textbf{27.9}\\ 
eggbox$^\dagger$ & 22.0    & 22.1      & 25.9 & 25.2     & 50.2           & -     & \textbf{52.4}\\ 
glue$^\dagger$   & 38.5    & 35.8      & 39.6 & 32.4     & 49.6           & -     & \textbf{53.8}\\ 
holepuncher      & 22.1    & 36.0      & 21.3 & 49.5     & 39.7           & -     & \textbf{54.2}\\ 
\hline
average          & 24.9    & 27.0      & 27.0 & 32.0     & 40.8           & 47.3  & \textbf{47.5}\\ 
\hline
\end{tabular}
\caption{\small{\textbf{Quantitative evaluation: ADD(-S) accuracy on Occlusion Linemod.} Baseline approaches: PoseCNN~\cite{XiangSNF18}, Oberweger et al.~\cite{Oberweger_2018}, Hu et al.~\cite{hu2019segmentation}, PVNet~\protect\cite{pvnet}, and DPOD~\cite{zakharov2019dpod}. Objects annotated with ($\dagger$) possess symmetric pose ambiguity. }}
\label{Figure:Qualitative:Occlusion}
\vspace{-0.1in}
\end{table}

As shown in Table~\ref{Figure:Qualitative:Lindmod}, Table~\ref{Figure:Qualitative:Occlusion}, and Figure~\ref{Figure:Pose:Regression:Result}, HybridPose leads to accurate pose estimation. On Linemod and Occlusion Linemod, HybridPose has an average ADD(-S) accuracy of 91.3 and 47.5, respectively. The result on Linemod outperforms all except one state-of-the-art approaches that regress poses from intermediate representations. The result on Occlusion-Linemod outperforms all state-of-the-art approaches.

\noindent\textbf{Baseline comparison on Linemod.} HybridPose outperforms PVNet~\cite{pvnet}, the backbone model we use to predict keypoints. The improvement is consistent across all object classes, which demonstrates clear advantage of using a hybrid as opposed to unitary intermediate representation. HybridPose shows competitive results against DPOD~\cite{zakharov2019dpod}, winning on six object classes. The advantage of DPOD comes from data augmentation and explicit modeling of dense correspondences between input and projected images, both of which cater to situations without object occlusion. A detailed analysis reveals that the classes of objects on which HybridPose exhibits sub-optimal performance are among the smallest objects in Linemod. It suggests that pixel-based descriptors used in our pipeline are limited by image resolution.

\noindent\textbf{Baseline comparison on Occlusion Linemod.} HybridPose outperforms all baselines. In terms of ADD(-S), our approach improves PVNet~\cite{pvnet} from 40.8 to 47.5, representing a 16.4\% enhancement, which clearly shows the advantage of HybridPose on occluded objects, where predictions of invisible keypoints can be noisy, and visible keypoints may not provide sufficient constraints for pose regression alone. HybridPose also outperforms DPOD, the state-of-the-art model on this dataset.

\noindent\textbf{Running time.} On a desktop with 16-core Intel(R) Xeon(R) E5-2637 CPU and GeForce GTX 1080 GPU, HybridPose takes 0.6 second to predict the intermediate representations, 0.4 second to regress the pose. Assuming a batch size of 30, this gives an an average processing speed of around 30 fps, enabling real-time analysis.

\subsection{Ablation Study}
\label{Section:Ablation:Study}

\begin{table}[t]
\centering
\footnotesize
\begin{tabular}{c|cc|cc|cc}
\hline
& \multicolumn{2}{c|}{keypoints} & \multicolumn{2}{c|}{keypoints + symmetries} &
\multicolumn{2}{c}{full model} \\ \cline{2-7}
& Rot. & Trans. & \quad Rot. & Trans. & Rot. & Trans. \\ \hline
ape            & 1.295\textdegree & 0.114 & \quad 1.295\textdegree & 0.114 & 1.241\textdegree & 0.079 \\
benchvise      & 1.295\textdegree & 0.038 & \quad 1.294\textdegree & 0.038 & 0.858\textdegree & 0.016 \\
cam            & 1.215\textdegree & 0.080 & \quad 1.215\textdegree & 0.080 & 1.133\textdegree & 0.043\\
can            & 1.305\textdegree & 0.052 & \quad 1.303\textdegree & 0.052 & 0.951\textdegree & 0.026 \\
cat            & 1.201\textdegree & 0.052 & \quad 1.201\textdegree & 0.052 & 1.050\textdegree & 0.041 \\
driller        & 1.267\textdegree & 0.040 & \quad 1.267\textdegree & 0.040 & 0.898\textdegree & 0.029\\
duck           & 1.738\textdegree & 0.099 & \quad 1.737\textdegree & 0.099 & 1.598\textdegree & 0.078 \\
eggbox         & 1.098\textdegree & 0.050 & \quad 1.098\textdegree & 0.050 & 1.008\textdegree & 0.044 \\
glue           & 1.440\textdegree & 0.071 & \quad 1.440\textdegree & 0.071 & 1.281\textdegree & 0.051 \\
holepuncher    & 1.434\textdegree & 0.062 & \quad 1.357\textdegree & 0.060 & 1.124\textdegree & 0.040 \\
iron           & 1.274\textdegree & 0.046 & \quad 1.274\textdegree & 0.046 & 1.058\textdegree & 0.019 \\
lamp           & 1.371\textdegree & 0.030 & \quad 1.371\textdegree & 0.030 & 0.903\textdegree & 0.022 \\
phone          & 1.678\textdegree & 0.055 & \quad 1.671\textdegree & 0.055 & 1.220\textdegree & 0.033 \\ \hline
mean           & 1.357\textdegree & 0.061 & \quad 1.350\textdegree & 0.061 & 1.104\textdegree & 0.040 \\
\hline
\end{tabular}
\caption{\textbf{Qualitative evaluation with different intermediate representations.} We report errors using two metrics: the median of absolute angular error in rotation, and the median of relative error in translation with respect to object diameter. }
\label{Figure:Quantitative:Ablation}
\vspace{-0.15in}
\end{table}

Table~\ref{Figure:Quantitative:Ablation} summarizes the performance of HybridPose using different predicted intermediate representations on the Linemod dataset. 

\noindent\textbf{With keypoints.} As a baseline approach, we estimate object poses by only utilizing keypoint information. This gives a mean absolute rotation error of 1.357\textdegree, and a mean relative translation error of 0.061.

\noindent\textbf{With keypoints and symmetry.}
Adding symmetry correspondences to keypoints leads to some performance gain in rotation. On the other hand, the translation error remains almost the same. One explanation is that symmetry correspondences only constrain two degrees of freedom in a total of three rotation parameters, and provide no constraint on translation parameters (see (\ref{Eq:3})).

\noindent\textbf{Full model.}
Adding edge vectors to keypoints and symmetry correspondences leads to salient performance gain in both rotation and translation estimations. One explanation is that edge vectors provide more constraints on both translation and rotation (see (\ref{Eq:2})).
Edge vectors provide more constraints on translation than keypoints as they represent adjacent keypoints displacement and provide gradient information for regression. Unlike symmetry correspondences, edge vectors constrain 3 degrees of freedom on rotation parameters which further boosts the performance of rotation estimation.

\section{Conclusions and Future Work}
\label{Section:Conclusions:Future:Work}

In this paper, we introduce HybridPose, a 6D pose estimation approach that utilizes keypoints, edge vectors, and symmetry correspondences. Experiments show that HybridPose enjoys real-time prediction and outperforms current state-of-the-art pose estimation approaches in accuracy. HybridPose is robust to occlusion. In the future, we plan to extend HybridPose to include more intermediate representations such as shape primitives, normals, and planar faces. Another possible direction is to enforce consistency across different representations in a similar way to~\cite{zhang2019path} as a self-supervision loss in network training. 

\section{Acknowledgement}
We would like to acknowledge the support of this research from NSF DMS-1700234, a Gift from Snap Research, and a hardware donation from NVIDIA.

\clearpage

\nocite{scikit-learn}
{\small
\bibliographystyle{ieee_fullname}
\bibliography{egbib}
}
\newpage
\includepdf[page=1]{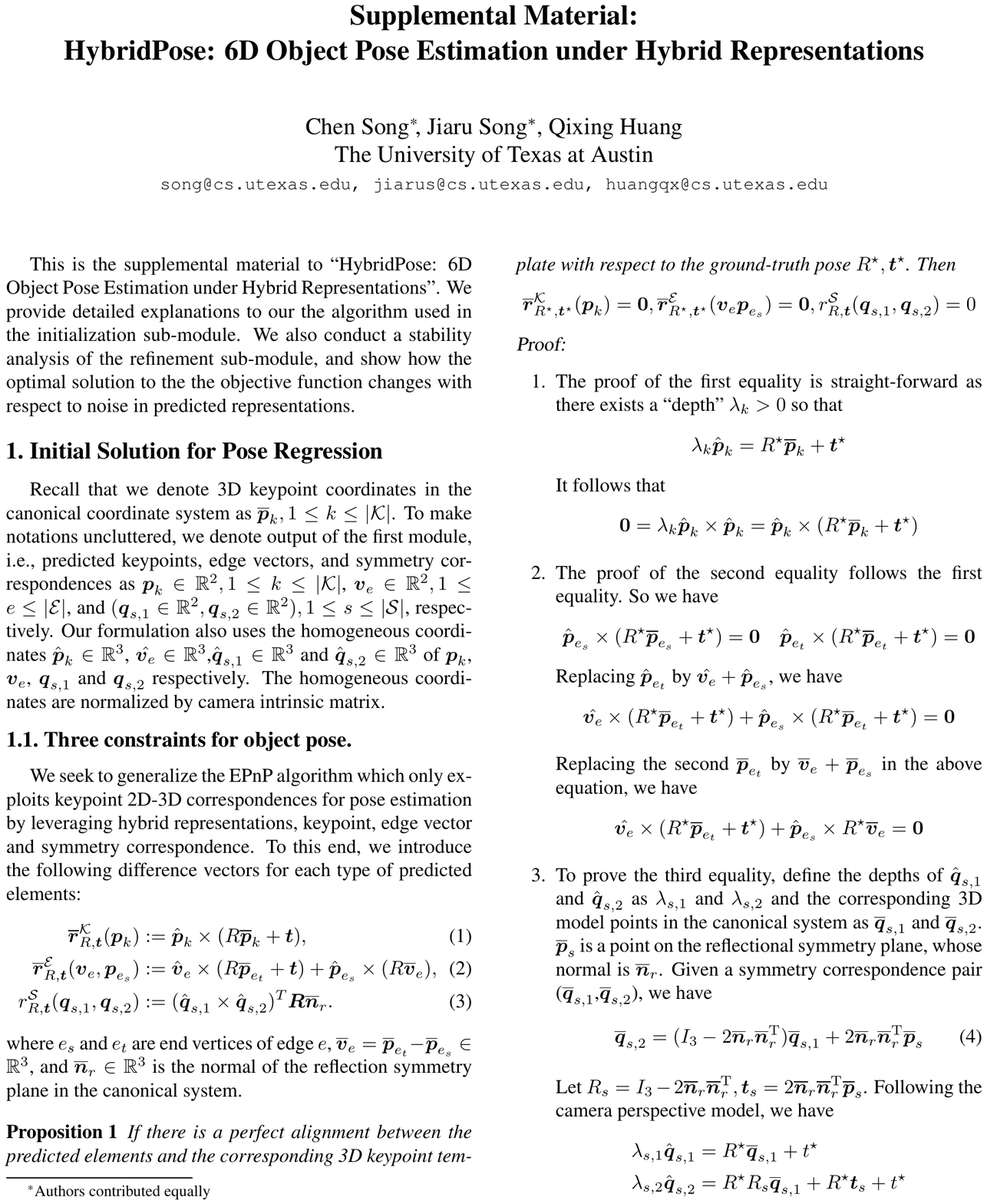}
\includepdf[page=2]{files/01176-supp.pdf}
\includepdf[page=3]{files/01176-supp.pdf}
\includepdf[page=4]{files/01176-supp.pdf}
\includepdf[page=5]{files/01176-supp.pdf}
\includepdf[page=6]{files/01176-supp.pdf}
\includepdf[page=7]{files/01176-supp.pdf}
\end{document}


\title{Supplemental Material: \\ HybridPose: 6D Object Pose Estimation under Hybrid Representations}

\author{Chen Song\thanks{Authors contributed equally} , Jiaru Song$^*$, Qixing Huang\\
The University of Texas at Austin\\
{\tt\small \hyperref[mailto:song@cs.utexas.edu]{song@cs.utexas.edu}, \hyperref[mailto:jiarus@cs.utexas.edu]{jiarus@cs.utexas.edu}, \hyperref[mailto:huangqx@cs.utexas.edu]{huangqx@cs.utexas.edu}}
}
\maketitle
\thispagestyle{empty}



This is the supplemental material to ``HybridPose: 6D Object Pose Estimation under Hybrid Representations''. We provide detailed explanations to our the algorithm used in the initialization sub-module. We also conduct a stability analysis of the refinement sub-module, and show how the optimal solution to the the objective function changes with respect to noise in predicted representations. 


\section{Initial Solution for Pose Regression}
Recall that we denote 3D keypoint coordinates in the canonical coordinate system as $\overline{\bs{p}}_k, 1\leq k \leq |\set{K}|$. To make notations uncluttered, we denote output of the first module, i.e., predicted keypoints, edge vectors, and symmetry correspondences as $\bs{p}_k\in \R^2, 1\leq k \leq |\set{K}|$, $\bs{v}_{e}\in \R^2, 1\leq e \leq |\set{E}|$, and  $(\bs{q}_{s,1}\in \R^2,\bs{q}_{s,2}\in \R^2), 1\leq s \leq |\set{S}|$, respectively. Our formulation also uses the homogeneous coordinates $\hat{\bs{p}}_k\in \R^3$, $\hat{\bs{v}_{e}}\in \R^3$,$\hat{\bs{q}}_{s,1}\in \R^3$ and $\hat{\bs{q}}_{s,2}\in \R^3$ of $\bs{p}_k$, $\bs{v}_{e}$, $\bs{q}_{s,1}$ and $\bs{q}_{s,2}$ respectively. The homogeneous coordinates are normalized by camera intrinsic matrix.
\subsection{Three constraints for object pose.}
We seek to generalize the EPnP algorithm which only exploits keypoint 2D-3D correspondences for pose estimation by leveraging hybrid representations, keypoint, edge vector and symmetry correspondence.  To this end, we introduce the following difference vectors for each type of predicted elements:
\begin{align}
\overline{\bs{r}}_{R, \bs{t}}^{\set{K}}(\bs{p}_k) & := \hat{\bs{p}}_k\times (R\overline{\bs{p}}_k+\bs{t}), \label{Eq:kpt} \\
\overline{\bs{r}}_{R,\bs{t}}^{\set{E}}(\bs{v}_e, \bs{p}_{e_s}) & := \hat{\bs{v}}_e\times (R\overline{\bs{p}}_{e_t}+\bs{t}) + \hat{\bs{p}}_{e_s}\times (R\overline{\bs{v}}_e), \label{Eq:edge} \\
r^{\set{S}}_{R,\bs{t}}(\bs{q}_{s,1},\bs{q}_{s,2}) & := (\hat{\bs{q}}_{s,1}\times \hat{\bs{q}}_{s,2})^T \bs{R}\overline{\bs{n}}_{r}.\label{Eq:sym_0}
\end{align}
where $e_s$ and $e_t$ are end vertices of edge $e$, $\overline{\bs{v}}_e = \overline{\bs{p}}_{e_t} - \overline{\bs{p}}_{e_s}\in \R^3$, and $\overline{\bs{n}}_r\in \R^3$ is the normal of the reflection symmetry plane in the canonical system. 

\begin{proposition}
If there is a perfect alignment between the predicted elements and the corresponding 3D keypoint template with respect to the ground-truth pose $R^{\star}, \bs{t}^{\star}$. Then 
$$
\overline{\bs{r}}_{R^{\star}, \bs{t}^{\star}}^{\set{K}}(\bs{p}_k) = \bs{0}, \\
\overline{\bs{r}}_{R^{\star},\bs{t}^{\star}}^{\set{E}}(\bs{v}_e\bs{p}_{e_s}) = \bs{0}, \\
r^{\set{S}}_{R,\bs{t}}(\bs{q}_{s,1},\bs{q}_{s,2}) = 0
$$
\end{proposition}

\noindent\textsl{Proof:} 
\begin{enumerate}
    \item The proof of the first equality is straight-forward as there exists a ``depth'' $\lambda_k > 0$ so that
$$
\lambda_k \hat{\bs{p}}_k = R^{\star}\overline{\bs{p}}_k + \bs{t}^{\star}
$$
It follows that 
$$
\bs{0} = \lambda_k \hat{\bs{p}}_k\times \hat{\bs{p}}_k = \hat{\bs{p}}_k\times (R^{\star}\overline{\bs{p}}_k + \bs{t}^{\star}) 
$$

\item 
The proof of the second equality follows the first equality. So we have 
$$
\hat{\bs{p}}_{e_s}\times (R^{\star}\overline{\bs{p}}_{e_s} + \bs{t}^{\star}) = \bs{0} \quad
\hat{\bs{p}}_{e_t}\times (R^{\star}\overline{\bs{p}}_{e_t} + \bs{t}^{\star}) = \bs{0} 
\label{Eq:sup1}
$$
Replacing $\hat{\bs{p}}_{e_t}$ by $ \hat{\bs{v}_e} + \hat{\bs{p}}_{e_s}$, we have
$$
\hat{\bs{v}_e}\times (R^{\star}\overline{\bs{p}}_{e_t} + \bs{t}^{\star}) + \hat{\bs{p}}_{e_s}\times (R^{\star}\overline{\bs{p}}_{e_t} + \bs{t}^{\star}) = \bs{0}
$$

Replacing the second $\overline{\bs{p}}_{e_t} $ by $\overline{\bs{v}}_e + \overline{\bs{p}}_{e_s}$ in the above equation, we have
$$
\hat{\bs{v}_e}\times (R^{\star}\overline{\bs{p}}_{e_t} + \bs{t}^{\star})  + \hat{\bs{p}}_{e_s}\times R^{\star}\overline{\bs{v}}_e = \bs{0}
$$
\item To prove the third equality, define the depths of $\hat{\bs{q}}_{s,1}$ and $ \hat{\bs{q}}_{s,2}$ as $\lambda_{s,1}$ and $\lambda_{s,2}$ and the corresponding 3D model points in the canonical system as $\overline{\bs{q}}_{s,1}$ and $\overline{\bs{q}}_{s,2}$. $\overline{\bs{p}}_s$ is a point on the reflectional symmetry plane, whose normal is $\overline{\bs{n}}_r$. Given a symmetry correspondence pair ($\overline{\bs{q}}_{s,1}$,$ \overline{\bs{q}}_{s,2}$), we have
\begin{equation}
    \overline{\bs{q}}_{s,2} = (I_3 - 2\overline{\bs{n}}_r\overline{\bs{n}}_r^{\mathrm{T}}) \overline{\bs{q}}_{s,1} + 2\overline{\bs{n}}_r\overline{\bs{n}}_r^{\mathrm{T}}\overline{\bs{p}}_s
    \label{Eq: sym}
\end{equation}

Let $R_s = I_3 - 2\overline{\bs{n}}_r\overline{\bs{n}}_r^{\mathrm{T}}, \bs{t}_s = 2\overline{\bs{n}}_r\overline{\bs{n}}_r^{\mathrm{T}}\overline{\bs{p}}_s$. Following the camera perspective model, we have
\begin{align*}
    \lambda_{s,1}\hat{\bs{q}}_{s,1} &= R^{\star}\overline{\bs{q}}_{s,1} + t^{\star}\\
    \lambda_{s,2}\hat{\bs{q}}_{s,2} &= R^{\star}R_s\overline{\bs{q}}_{s,1} + R^{\star}\bs{t}_s+ t^{\star}
\end{align*}
Subtracting these two equations, we have
$$
 \lambda_{s,2}\hat{\bs{q}}_{s,2} - \lambda_{s,1}\hat{\bs{q}}_{s,1} = R^{\star}(R_s\overline{\bs{q}}_{s,1} + \bs{t}_s - \overline{\bs{q}}_{s,1})
 \label{Eq:sym1}
$$
Left multiply both sides of the equation by $\hat{\bs{q}}_{s,2}\times$ yields
\begin{equation}
- \lambda_{s,1}\hat{\bs{q}}_{s,2}\times \hat{\bs{q}}_{s,1} = \hat{\bs{q}}_{s,2}\times[R^{\star}(R_s\overline{\bs{q}}_{s,1} + \bs{t}_s - \overline{\bs{q}}_{s,1})]
 \label{Eq:sym1}
\end{equation}
Geometrically, (\ref{Eq:sym1}) reveals that $\hat{\bs{q}}_{s,2}\times \hat{\bs{q}}_{s,1}$ is perpendicular to the plane with span of $\{{\bs{q}}_{s,2}, R^{\star}(R_s\overline{\bs{q}}_{s,1} + \bs{t}_s - \overline{\bs{q}}_{s,1})\}$, thus we have 
\begin{align*}
    (\hat{\bs{q}}_{s,2}\times \hat{\bs{q}}_{s,1})^{\mathrm{T}}R^{\star}(R_s\overline{\bs{q}}_{s,1} + \bs{t}_s - \overline{\bs{q}}_{s,1}) =\\
    2(\overline{\bs{n}}_r^{\mathrm{T}}(\overline{\bs{p}}_s - \overline{\bs{q}}_{s,1}))(\hat{\bs{q}}_{s,2}\times \hat{\bs{q}}_{s,1})^{\mathrm{T}}R^{\star}\overline{\bs{n}}_r = 0
\end{align*}
Since $2(\overline{\bs{n}}_r^{\mathrm{T}}(\overline{\bs{p}}_s - \overline{\bs{q}}_{s,1}))$ is a non-zero scalar, we can delete this term and finally get
$$
(\hat{\bs{q}}_{s,2}\times \hat{\bs{q}}_{s,1})^{\mathrm{T}}R^{\star}\overline{\bs{n}}_r = 0
$$
\end{enumerate}
\subsection{Pose solution in eigenvector space.}
\label{subsec:1.2}
A nice feature shared by (\ref{Eq:kpt}), (\ref{Eq:edge}) and (\ref{Eq:sym_0}) is that all constraints are linear in the elements of $R$ and $\bs{t}$. This allows us to derive a closed-form solution of $R$ and $\bs{t}$ in the affine transformation space. Specifically, we can define $\bs{x} = (\bs{r}_1^{\mathrm{T}},\bs{r}_2^{\mathrm{T}},\bs{r}_3^{\mathrm{T}},\bs{t}^{\mathrm{T}})_{12\times1}^{\mathrm{T}}$ as a vector that contains rotation and translation parameters in affine space. Expanding constraint (\ref{Eq:kpt}) and constraint (\ref{Eq:edge}) yields three linear equations for each predicted element respectively for $\bs{x}$, and expanding constraint (\ref{Eq:sym_0}) yields one linear equation. By concatenating all linear equations of predicted elements together, we can generate a linear system of the form $A\bs{x} = \bs{0}$, where $A$ is matrix and its dimension is $(3|\set{K}| + 3|\set{E}| + |\set{S}|)\times 12$. 

To model the relative importance among keypoints, edge vectors, and symmetry correspondences, we rescale (\ref{Eq:edge}) and (\ref{Eq:sym_0}) by hyper-parameters $\alpha_{E}$ and $\alpha_{S}$, respectively, to generate $A$. As discussed in the body of this paper, we calculate $\alpha_{E}$ and $\alpha_{S}$ by solving an optimization problem using finite-difference and back-track line search.

Then following EPnP~\cite{lepetit2009epnp}, we compute $\bs{x}$ as 
\begin{equation}
    \bs{x} = \sum_{i = 1}^N\gamma_i\bs{v}_i 
\end{equation}
where $\bs{v}_i$ is the $i^{\mathrm{th}}$ smallest right singular vector of $A$. Ideally, when predicted elements are noise-free, $N = 1$ with $\bs{x} = \bs{v}_1$ is an optimal solution. However, this strategy performs poorly given noisy predictions. Same as EPnP~\cite{lepetit2009epnp}, we choose $N = 4$. 

\subsection{Optimize a good linear combination.}
To compute the optimal $\bs{x}$, we optimize latent variables $\gamma_i$ and the rotation matrix $R$ with following objective function:
\begin{equation}
\underset{R\in \R^{3\times 3},\gamma_i}{\min} \| \sum_{i = 1}^4\gamma_i R_i - R\|_{\set{F}}^2
\label{Eq:Init:objective}
\end{equation}
where $R_i \in \R^{3\times 3}$ is reshaped from the first $9$ elements of $\bs{v}_i$. We solve this optimization problem with the following alternating procedure:
\begin{enumerate}
    \item Fix $\gamma_i$ and solve for $R$ by SVD. i.e. $R = U\mathrm{diag}(1,1,1)V^{\mathrm{T}}$ given $\sum_{i = 1}^4\gamma_i R_i = U\Sigma V^{\mathrm{T}}$\footnote{If $\mathrm{det}(R) < 0$ we enforce $\mathrm{det}(R) > 0$ by defining $R = U\mathrm{diag}(1,1,-1)V^{\mathrm{T}}$. },
    \item Fix $R$ and solve for $\gamma_i$'s by optimizing a linear system $\sum_{i = 1}^4\gamma_i R_i = R$ in an element-wise manner.
\end{enumerate}

To initialize $\gamma_i$'s for the above optimization problem, we calculate $\gamma_i$ with $i = 1...3$ by enforcing that $\sum_{i = 1}^3\gamma_i R_i$ is an orthogonal matrix\footnote{The reason of initializing 3 $\gamma_i$'s is that (\ref{gamma_init}) is unable to provide enough linear constraints for 4 $\gamma_i$'s and this initialization ensures the convergence of optimization.}:
\begin{equation}
    (\sum_{i = 1}^3\gamma_i R_i)^{\mathrm{T}}\sum_{i = 1}^3\gamma_i R_i = I_3
    \label{gamma_init}
\end{equation}

Since $I_3$ is a symmetric matrix, expanding (\ref{gamma_init}) yields 6 nonlinear constraints for $\bs{\gamma} = (\gamma_1,\gamma_2,\gamma_3)^{\mathrm{T}}$, which is however uneasy to solve. We then define a new vector $\bs{y} = (y_1,y_2,y_3,y_4,y_5,y_6)^{\mathrm{T}} = (\gamma_1^2,\gamma_1\gamma_2,\gamma_1\gamma_3,\gamma_2^2,\gamma_2\gamma_3,\gamma_3^2)^{\mathrm{T}}$ and form a linear system $C\bs{y} = \bs{z}$ which has the unique solution with $\bs{z}$ generated from $I_{3}$. Afterwards, it is easy to recover $\gamma_i$ from $\bs{y}$ and optimize from initialized $\gamma_i$ alone with $\gamma_4 = 0$. 

After optimization, we again apply SVD to project $\sum_{i = 1}^4\gamma_i R_i$ onto the space of SO(3), i.e., $R^{init} = U\mathrm{diag}(1,1,1)V^{\mathrm{T}}$ and enforce $\mathrm{det}(R^{init}) > 0$ where $R^{init} = U\Sigma V^T$. Leveraging $A\bs{x} = 0$ defined in section (\ref{subsec:1.2}), the corresponding translation $\bs{t}^{init}$ is
\begin{equation}
\bs{t}^{\init}=-(A_2^{\mathrm{T}}A_2)^{-1}A_2^{\mathrm{T}}A_1\bs{r}^{init}
\label{Eq:Init:Output}
\end{equation}
where $A_1 = A_{[:,1:9]}$, $A_2 = A_{[:,10:12]}$, $\bs{r}^{init}_{9\times1}$ is reshaped from $R^{init}$.

\section{Stability Analysis for Pose Refinement}

In this section, we provide a local stability analysis of the pose regression procedure, which amounts to solving the following optimization problem:
\begin{align}
\min\limits_{R,\bs{t}} & \ \sum\limits_{k=1}^{|\set{K}|} \rho(\|\bs{r}^{\set{K}}_{R,\bs{t}}(\bs{p}_k)\|,\beta_{\set{K}}) \|\bs{r}^{\set{K}}_{R,\bs{t}}(\bs{p}_k)\|_{\Sigma_k}^2 \nonumber \\
& + \frac{|\set{K}|}{|\set{E}|}\sum\limits_{e=1}^{|\set{E}|}
\rho(\|\bs{r}^{\set{E}}_{R,\bs{t}}(\bs{v}_e)\|,\beta_{\set{E}}) \|\bs{r}^{\set{E}}_{R,\bs{t}}(\bs{v}_e)\|_{\Sigma_e}^2
\nonumber \\
& \ + \frac{|\set{K}|}{|\set{S}|}\sum\limits_{s=1}^{|\set{S}|}\rho(r^{\set{S}}_{R,\bs{t}}(\bs{q}_{s,1},\bs{q}_{s,2}), \beta_{\set{S}})r^{\set{S}}_{R,\bs{t}}(\bs{q}_{s,1},\bs{q}_{s,2})^2
\label{Eq:Pose2}
\end{align}

When predictions are accurate, then the optimal solution of the objective function described above should recover the underlying ground-truth. However, when the predictions possess noise, then the optimal object pose can drift from the underlying ground-truth. Our focus is local analysis, which seeks to understand the interplay between different objective terms defined by keypoints, edge vectors, and symmetry correspondences. Therefore, we assume the noise level of the input is small, and the perturbation of the output is well captured by low-order Taylor expansion of the output. 

Our goal is to characterize the relation between the variance of the input noise and the variance of the output pose. We show that incorporating edge vectors and symmetry correspondences generally help to reduce the variance of the output. 

The remainder of this section is organized as follows. In Section~\ref{Section:Local:Stability}, we provide a local stability analysis framework for regression problems. In Section~\ref{Section:Structure:Pose:Stability}, we describe the structure of the pose regression and apply this framework to provide a preliminary analysis of the stability of pose regression. In Section~\ref{Section:Specific:Example}, we provide further analysis on a specific example, which indicates the interactions among keypoints, edge vectors, and symmetry correspondences. Finally, Section~\ref{Section:Proof:Propositions} provide proofs of the propositions in this analysis.

\subsection{Local Stability Analysis Framework}
\label{Section:Local:Stability}

We begin with a general result regarding an optimization problem of the following form
\begin{equation}
\bs{x}^{\star}(\bs{y}):= \underset{\bs{y}}{\textup{argmin}}\ f(\bs{x},\bs{y}).
\label{Eq:F:Opt:Def}
\end{equation}
In the context of this paper, $\bs{y}$ encodes the noise associated with the predictions, i.e., keypoints, edge vectors, and symmetry correspondences.  $\bs{x}\in \R^6$ provides a local parameterization of the output, i.e., the object pose. The specific expressions of $\bs{y}$ and $\bs{x}$ will be described in Section~\ref{Section:Structure:Pose:Stability}.

Without losing generality, we further assume that $f$ satisfies the following assumptions (which are valid in the context of this paper):
\begin{itemize}
\item $f(\bs{x},\bs{y})\geq 0$. Moreover, $f(\bs{x},\bs{y}) = 0$ if and only if $\bs{x} = 0$ and $\bs{y} = 0$. This means $\bs{x}^{\star}(\bs{0}) = \bs{0}$, and $(\bs{0},\bs{0})$ is the strict global optimal solution. 

\item $f$ is smooth and at least $C^3$ continuous.

\item The following Hessian matrix is positive definite in some local neighborhood of $(\bs{0},\bs{0})$:
$$
\left[
\begin{array}{cc}
\frac{\partial^2 f}{\partial^2 \bs{x}} & \frac{\partial^2 f}{\partial \bs{x} \partial \bs{y}} \\
\frac{\partial^2 f}{\partial \bs{x} \partial \bs{y}} & \frac{\partial^2 f}{\partial^2 \bs{y}} 
\end{array}
\right].
$$ 
\end{itemize}

Our analysis will utilize the following partial derivative of $\bs{x}^{\star}$ with respect to $\bs{y}$.
\begin{proposition}
Under the assumptions described above, $\bs{x}^{\star}(\bs{y})$ is unique in the local neighborhood of $\bs{0}$, and
\begin{equation}
\frac{\partial \bs{x}^{\star}}{\partial \bs{y}}(\bs{y}):= - \big(\frac{\partial^2 f}{\partial^2 \bs{x}}(\bs{x}^{\star}(\bs{y}),\bs{y})\big)^{-1}\frac{\partial^2 f}{\partial \bs{x}\partial \bs{y}}(\bs{x}^{\star}(\bs{y}),\bs{y}).
\label{Eq:LSA:2}
\end{equation}
\label{Prop:GIMF}
\end{proposition}
\noindent\textbf{Proof.} See Section~\ref{Proof:Prop:GIMF}. $\square$

Since we are interested in local stability analysis, we assume the magnitude of $\bs{y}$ is small. Thus,
\begin{equation}
\bs{x}^{\star}(\bs{y}) \approx \frac{\partial \bs{x}^{\star}}{\partial \bs{y}}(\bs{0})\cdot \bs{y}.
\label{Eq:LSA:5}
\end{equation}
If we further assume $\bs{y}$ follows some random distribution whose variance matrix if $\textup{Var}(\bs{y})$.
Then the variance of the output $\bs{x}^{\star}$ is given by
\begin{align}
& \textup{Var}(\bs{x}^{\star}(\bs{y})) \nonumber \\
\approx & \big(\frac{\partial^2 f}{\partial^2 \bs{x}}\big)^{-1}\cdot \frac{\partial^2 f}{\partial \bs{x}\partial \bs{y}}\cdot \textup{Var}(\bs{y})(\frac{\partial^2 f}{\partial \bs{x}\partial \bs{y}})^T\big(\frac{\partial^2 f}{\partial^2 \bs{x}}\big)^{-1}.
\label{Eq:Pose:Variance}
\end{align}

Note that in our problem, $f$ consists of non-linear least squares, i.e., 
\begin{equation}
f = \sum\limits \frac{\beta_{i,1}^2\cdot \|\bs{r}_i\|_{\Sigma_i}^2}{\beta_{i,2}^2 + \|\bs{r}_i\|^2}.
\label{Eq:F:Form}
\end{equation}
The following proposition characterizes how to compute $\frac{\partial^2 f}{\partial^2 \bs{x}}$ and $\frac{\partial^2 f}{\partial \bs{x} \partial \bs{y}}$.
\begin{proposition}
Under the expression described in (\ref{Eq:F:Form}), the second-order derivatives $\frac{\partial^2 f}{\partial^2 \bs{x}}$ and $\frac{\partial^2 f}{\partial \bs{x} \partial \bs{y}}$ at $(\bs{0},\bs{0})$ are given by 
\begin{align}
\frac{\partial^2 f}{\partial^2 \bs{x}} &= \sum\limits_{i} \frac{\beta_{i,1}^2}{\beta_{i,2}^2}\frac{\partial \bs{r}_i}{\partial \bs{x}}\Sigma_i{\frac{\partial \bs{r}_i}{\partial \bs{x}}}^T \\
\frac{\partial^2 f}{\partial \bs{x} \partial \bs{y}} &= \sum\limits_{i} \frac{\beta_{i,1}^2}{\beta_{i,2}^2}\frac{\partial \bs{r}_i}{\partial \bs{x}}\Sigma_i{\frac{\partial \bs{r}_i}{\partial \bs{y}}}^T
\end{align}
\label{Prop:GNLS}
\end{proposition}

\noindent\textbf{Proof.} See Section~\ref{Proof:Prop:GNLS}. $\square$

\subsection{Structure of Pose Stability}
\label{Section:Structure:Pose:Stability}

We begin by rephrasing the pose-regression problem described in the main paper. 

\noindent\textbf{Ground-truth setup.} We use the same definition of variables as that in full paper. Recall that $\overline{\bs{p}}_k$ is coordinates of keypoint in canonical system. Let $R^{\gt}$ and $\bs{t}^{\gt}$ be the ground-truth pose. Then the ground-truth 3D location of $\overline{\bs{p}}_k$ in the camera coordinate system is
$$
\overline{\bs{p}}_k^{\gt} = R^{\gt}\overline{\bs{p}}_k + \bs{t}^{\gt},\qquad 1\leq k \leq |\set{K}|.
$$
Let $\overline{\bs{p}}_k^{\gt} = (p_{k,x}^{\gt, 3D},p_{k,y}^{\gt,3D},p_{k,z}^{\gt,3D})^T$. Then the ground-truth image coordinates of the projected keypoint $\bs{p}_k^{\gt}\in\R^2$ is given by 
$$
\bs{p}_{k}^{\gt} = (\frac{p_{k,x}^{\gt, 3D}}{p_{k,z}^{\gt, 3D}},\frac{p_{k,y}^{\gt, 3D}}{p_{k,z}^{\gt, 3D}})^T = (p_{k,x}^{\gt},p_{k,y}^{\gt})^T.
$$
Likewise, recall $(\overline{\bs{q}}_{s1},\overline{\bs{q}}_{s2})$ are symmetry correspondence in the world coordinate system, and let 
\begin{align*}
\overline{\bs{q}}_{s1}^{\gt} & := R^{\gt}\overline{\bs{q}}_{s1} + \bs{t}^{\gt} \\
\overline{\bs{q}}_{s2}^{\gt} & := R^{\gt}\overline{\bs{q}}_{s2} + \bs{t}^{\gt} \\
\end{align*}
denote the transformed points in the camera coordinate system, where $\overline{\bs{q}}_{si}^{\gt} = (q_{si,x}^{\gt, 3D},q_{si,y}^{\gt,3D},q_{si,z}^{\gt,3D})^T$. So the image coordinates of each symmetry correspondence are given by 
$$
\bs{q}_{s1}^{\gt} = (\frac{q_{s1,x}^{\gt,3D}}{q_{s1,z}^{\gt,3D}},\frac{q_{s1,y}^{\gt,3D}}{q_{s1,z}^{\gt,3D}})^T,\quad \bs{q}_{s2}^{\gt} =(\frac{q_{s2,x}^{\gt,3D}}{q_{s2,z}^{\gt,3D}},\frac{q_{s2,y}^{\gt,3D}}{q_{s2,z}^{\gt,3D}})^T.
$$

\noindent\textbf{Noise model.} we proceed to describe the noise model used in the stability analysis. In this analysis, we assume each input keypoint is perturbed from the ground-truth location by $\bs{y}_{k} = (y_{k,x},y_{k,y})^T$, i.e.,
$$
\bs{p}_k =\bs{p}_k^{\gt} + \bs{y}_k.
$$
Likewise, we assume each input edge vector is perturbed from the ground-truth edge vector by $\bs{y}_e = (y_{e,x},y_{e,y})^T$, i.e.,
$$
\bs{v}_e = \bs{p}_{e_s}^{\gt} - \bs{p}_{e_t}^{\gt} + \bs{y}_e.
$$
Finally, for symmetry correspondences, we assume that $\bs{q}_{s1}$ is not perturbed), and $\bs{q}_{s2}$ is perturbed by $\bs{y}_s = (y_{s,x},y_{s,y})^T$, i.e.,
$$
\bs{q}_{s1} = \bs{q}_{s1}^{\gt},\quad \bs{q}_{s2} = \bs{q}_{s2}^{\gt}+ \bs{y}_s.
$$

\noindent\textbf{Local parameterization.} We parameterize the 6D object pose locally using exponential map with coefficients $(\bs{c}\in \R^3,\overline{\bs{c}}\in \R^3)$, i.e.,
$$
R = \mathrm{exp}(\bs{c}\times)\cdot R^{\gt}, \quad \bs{t} = \bs{t}^{\gt} + \overline{\bs{c}}.
$$
Note this parameterization is quite standard for rigid transformations.

Now consider the three terms used in pose regression\footnote{For convenience, we negate both $\bs{r}_k^{\set{K}}$ and $\bs{r}_e^{\set{E}}$ defined in the body of this paper.}:
\begin{align*}
\bs{r}_k^{\set{K}} &:= \bs{p}_k - \set{P}_{R,\bs{t}}(\overline{\bs{p}}_k), \\
\bs{r}_e^{\set{E}} &:= \bs{v}_e - (\set{P}_{R,\bs{t}}(\overline{\bs{p}}_{e_s})-\set{P}_{R,\bs{t}}(\overline{\bs{p}}_{e_t})), \\
r_s^{\set{S}} &:= (\hat{\bs{q}}_{s,1}\times \hat{\bs{q}}_{s,2})^T R \overline{\bs{n}}_r.
\end{align*}

The following proposition characterizes the derivatives between each term and the parameters of the noise model and the parameters of the local parameterization.

\begin{proposition}
Define 
\begin{align*}
J_{k,\bs{c}} &= \left(
\begin{array}{ccc}
-p_{k,x}^{\gt}p_{k,y}^{\gt} & 1 + {p_{k,x}^{\gt}}^2 & -p_{k,y}^{\gt} \\
-1 - {p_{k,y}^{\gt}}^2 & p_{k,x}^{\gt}p_{k,y}^{\gt} & p_{k,x}^{\gt}
\end{array}
\right)\\
J_{k,\overline{\bs{c}}} &= \frac{1}{p_{k,z}^{\gt,3D}}\left(
\begin{array}{ccc}
1 & 0 & -p_{k,x}^{\gt} \\
0 & 1 & -p_{k,y}^{\gt}
\end{array}
\right)\\
\end{align*}
The derivatives of $\bs{r}_{R,\bs{t}}^{\set{K}}(\bs{p}_k) = \bs{r}^{\set{K}}_k$ are given by
\begin{align*}
\left(\frac{\partial \bs{r}^{\set{K}}_k}{\partial \bs{y}_{k}},\frac{\partial \bs{r}^{\set{K}}_k}{\partial \bs{c}},\frac{\partial \bs{r}^{\set{K}}_k}{\partial \overline{\bs{c}}}\right) &= (I_2,- J_{k,\bs{c}},- J_{k,\overline{\bs{c}}})
\end{align*}
The derivatives of $\bs{r}_{R,\bs{t}}^{\set{E}}(\bs{v}_e) = \bs{r}_e^{\set{E}}$ are given by
\begin{align*}
\left(\frac{\partial \bs{r}_e^{\set{E}}}{\partial \bs{y}_{e}},\frac{\partial \bs{r}_e^{\set{E}}}{\partial \bs{c}},\frac{\partial \bs{r}_e^{\set{E}}}{\partial \overline{\bs{c}}}\right)&= (I_2, J_{e_t,\bs{c}}-J_{e_s,\bs{c}}, J_{e_t,\overline{\bs{c}}}-J_{e_s,\overline{\bs{c}}}).
\end{align*}
Moreover, the derivatives of $r_s^{S}$ are given by
\begin{align}
\frac{\partial r_s^{S}}{\partial \bs{y}_s} & = \left(
\begin{array}{c}
n_y^{\gt} - n_z^{\gt}q_{s1,y}^{\gt} \\
-(n_x^{\gt} - n_z^{\gt} q_{s1,x}^{\gt})
\end{array}
\right)^T, \label{Eq:LSA:6}\\
\frac{\partial r_s^{S}}{\partial \bs{c}} & = \big(\overline{\bs{n}}_r\times (\hat{\bs{q}}_{s1}^{\gt}\times \hat{\bs{q}}_{s2}^{\gt})\big)^T. \label{Eq:LSA:7}
\end{align}
where $\overline{\bs{n}}^{gt} = R\overline{\bs{n}}_r = (n_x^{\gt}, n_y^{\gt}, n_z^{\gt})^T$, $\hat{\bs{q}}_{si}^{\gt}$ is homogeneous coordinate of $\bs{q}_{si}^{\gt}$ normalized by camera intrinsic matrix.
\label{Prop:Jacobi:1}
\end{proposition}
\noindent\textbf{Proof.} See Section~\ref{Proof:Jacobi:1}. $\square$


Let $\bs{y}$ collect all the random variables in a vector. Let $J_{\set{K}}$, $J_{\set{E}}$, and $J_{\set{S}}$ collect the Jacobi matrices for the predicted elements under each type in its column. Note that the size of $J_{\set{S}}$ is $3\times3$ according to the derivations above. To facilitate the definition below, we reshape $J_{\set{S}}$ as a $6\times6$ matrix by placing original elements to the upper-left corner, and zeros to elsewhere. Denote $\beta_{\set{E}}$ and $\beta_{\set{S}}$ as the weight in front of each term (without loss of generality, we set $\beta_{\set{K}} = 1$). Then the variance matrix $\textup{Var}(\bs{c},\overline{\bs{c}})$ can be approximated by
\begin{equation}
\textup{Var}(\bs{c},\overline{\bs{c}}) \approx A^{-1}B \textup{Var}(\bs{y})B^T A^{-1}
\label{Eq:Var:Mat}
\end{equation}
where
\begin{align*}
A:= & J_{\set{K}} J_{\set{K}}^T + \beta_{\set{E}}J_{\set{E}} J_{\set{E}}^T + \beta_{\set{S}}J_{\set{S}} J_{\set{S}}^T \\
B:= & (J_{\set{K}},\beta_{\set{E}}J_{\set{E}},\beta_{\set{S}}J_{\set{S}})
\end{align*}
It we consider $A^{-1}B \textup{Var}(\bs{y})B^T A^{-1}$ as a function of $\beta_{\set{E}}$ and $\beta_{\set{S}}$ and compute its derivatives at $\beta_{\set{E}} = 0$ and $\beta_{\set{S}} = 0$, we obtain
\begin{align*}
& \frac{\partial A^{-1}B \textup{Var}(\bs{y})B^T A^{-1}}{\partial \beta_{\set{E}}} \\
:= & A^{-1} \big(J_{\set{E}}\Sigma_{\set{E}\set{K}} J_{\set{K}}^T + J_{\set{K}}\Sigma_{\set{K}\set{E}} J_{\set{E}}^T \\
& - J_{\set{E}}J_{\set{E}}^T A^{-1}J_{\set{K}}\Sigma_{\set{K}\set{K}} J_{\set{K}}^T-J_{\set{K}}\Sigma_{\set{K}\set{K}}J_{\set{K}}^T A^{-1}J_{\set{E}}J_{\set{E}}^T\big) A^{-1}.
\end{align*}
where $\Sigma_{\set{K}\set{K}}$ and $\Sigma_{\set{E}\set{K}}$ are the corresponding components in $\textup{Var}(\bs{y})$. This means whenever
\begin{align}
&J_{\set{E}}\Sigma_{\set{E}\set{K}} J_{\set{K}}^T + J_{\set{K}}\Sigma_{\set{K}\set{E}} J_{\set{E}}^T  \nonumber \\
\prec &J_{\set{E}}J_{\set{E}}^T A^{-1}J_{\set{K}}\Sigma_{\set{K}\set{K}} J_{\set{K}}^T+J_{\set{K}}\Sigma_{\set{K}\set{K}}J_{\set{K}}^T A^{-1}J_{\set{E}}J_{\set{E}}^T,
\label{Eq:Cond:Positive:2}
\end{align}
increasing the value of $\beta_{\set{E}}$ from zero is guaranteed to obtain a positive reduction in the variance matrix (in terms of both the trace-norm and the spectral-norm). 

(\ref{Eq:Cond:Positive:2}) is satisfied when $\Sigma_{\set{K}\set{K}} = I$ and $\Sigma_{\set{K}\set{E}} = 0$. In general, when $\bs{y}_k$ and $\bs{y}_{e}$ are uncorrelated, then it is likely that increasing its value can lead to reduction in the output variance matrix. 

A very similar argument can be applied to $\beta_{\set{S}}$, and we omit the details for brevity.

\subsection{An Example}
\label{Section:Specific:Example}

We proceed to provide an example that explicitly shows how the variance of $\textup{Var}([\bs{c},\overline{\bs{c}}])$ is reduced by incorporating edge vectors and symmetry correspondences. To this end, we consider a simple object that is given by a square, whose normal direction is along the z-axis in the camera coordinate system. We assume this square object has eight keypoints, whose $z$ coordinates are all $1$, i.e., $p_{k,z}^{\gt,3D} = 1, 1\leq k \leq 8$. Their $x$ and $y$ image coordinates are:
\begin{align*}
\bs{p}_{1}^{\gt} = (\delta, \delta), \quad \bs{p}_{2}^{\gt} = (\delta, 0), \quad \bs{p}_{3}^{\gt} = (\delta, -\delta), \\
\quad \bs{p}_{4}^{\gt} = (0, \delta), \quad \bs{p}_{5}^{\gt} = (0, -\delta), \quad \bs{p}_{6}^{\gt} = (-\delta, \delta),\\
\bs{p}_{7}^{\gt} = (-\delta, 0), \quad \bs{p}_{8}^{\gt} = (-\delta, -\delta)
\end{align*}
Moreover, assume that the normal to the reflection plane is $(1,0,0)$. The ground-truth symmetry correspondences are dense, and they are in the form of $(x,y)$ and $(-x,y)$, where $0\leq x \leq 1, -1 \leq y \leq 1$. 

With this setup and after simple calculations, we have 
\begin{align*}
H_{\set{K}}:=& \frac{1}{8}\sum\limits_{k=1}^{8} J_k^T J_k \\
= & \left(
\begin{array}{cccccc}
c_1(\delta) & 0 & 0 & 0 & -c_2(\delta) & 0 \\
0 &c_1(\delta) & 0 & c_2(\delta) & 0 & 0 \\
0 & 0 & \frac{12\delta^2}{8} & 0 & 0 & 0 \\
0 & c_2(\delta) & 0 & 1 & 0 & 0 \\
-c_2(\delta) & 0 & 0 & 0 & 1 & 0\\
0 & 0 & 0 & 0 & 0 &\frac{12\delta^2}{8}
\end{array}
\right)
\end{align*}
where $c_1(\delta) = \frac{8 + 12 \delta^2 + 10\delta^4}{8}$, and $c_2(\delta) = \frac{8+6\delta^2}{8}$.

Likewise, we have
\begin{align*}
H_{\set{E}}:=& \frac{1}{28}\sum\limits_{e=1}^{28} J_e^T J_e \\
= & \left(
\begin{array}{cccccc}
\frac{11}{7}\delta^4 & 0 & 0 & 0 & 0 & 0 \\
0 &\frac{11}{7}\delta^4 & 0 & 0 & 0 & 0 \\
0 & 0 & \frac{24}{7}\delta^2 & 0 & 0 & 0 \\
0 & 0 & 0 & 0 & 0 & 0 \\
0 & 0 & 0 & 0 & 0 & 0\\
0 & 0 & 0 & 0 & 0 &\frac{24}{7}\delta^2
\end{array}
\right)
\end{align*}

Finally, we have
\begin{align*}
H_S := & [\int_{0}^1(\int_{-1}^{1}J_s^T J_s dy) dx,0;0,0] \\
= &\left(
\begin{array}{cccccc}
0 & 0 & 0 & 0 & 0 & 0\\
0 & \frac{4}{3}\delta^2 & 0 & 0 & 0 &0\\
0 & 0 & \frac{4}{9}\delta^4 & 0 & 0 & 0\\
0 & 0 & 0 & 0 & 0 & 0 \\
0 & 0 & 0 & 0 & 0 & 0 \\
0 & 0 & 0 & 0 & 0 & 0 
\end{array}
\right)
\end{align*}

We proceed to assume the following noise model for the input:
\begin{align*}
\textup{Var}(\bs{y}_k) = \sigma_{\set{K}}^2 I_2,\quad \textup{Var}(\bs{y}_e) = \sigma_{\set{E}}^2 I_2, \quad \textup{Var}(y_s) = \sigma_{\set{S}}^2.    
\end{align*}
In other words, noises in different predictions are independent. 

Applying Prop.~\ref{Proof:Prop:GNLS}, we have that
\begin{align}
& \textup{Var}([\bs{c},\overline{\bs{c}}]) \nonumber\\
\approx&  (H_{\set{K}}+\lambda H_{\set{E}}+\mu H_{\set{S}})^{-1}\cdot \nonumber \\
&(\sigma_{\set{K}}^2 H_{\set{K}}+\lambda^2\sigma_{\set{E}}^2 H_{\set{E}}+\mu^2 \sigma_{\set{S}}^2 H_{\set{S}})\cdot \nonumber\\
&(H_{\set{K}}+\lambda H_{\set{E}}+\mu H_{\set{S}})^{-1} \nonumber \\
= & \left(
\begin{array}{cccccc}
a_1 & 0 & 0 & 0 & a_2 & 0 \\
0 & a_3 & 0 & a_4 & 0 & 0 \\
0 & 0 & a_5 & 0 & 0 & 0 \\
0 & a_4 & 0 & a_6 & 0 & 0 \\
a_2 & 0 & 0 & 0 & a_7 & 0 \\
0 & 0 & 0 & 0 & 0 & a_8 
\end{array}
\right)
\label{Eq:LSA:10}
\end{align}
$a_i, 1\leq i \leq 8$ are functions of $\delta, \beta_{\set{E}}, \beta_{\set{S}}$ and $\sigma_{\set{K}}^2, \sigma_{\set{E}}^2, \sigma_{\set{S}}^2$. For simplify, we only analyze  $a_8$, which is
\begin{align*}
a_8 & = \frac{\sigma_{\set{K}}^2\frac{12}{8}\delta^2 + \beta_{\set{E}}^2 \sigma_{\set{E}}^2 \frac{24}{7}\delta^2}{\big(\frac{12}{8}\delta^2 + \frac{24}{7}\beta_{\set{E}}\delta^2\big)^2}
\end{align*}
It is easy to check that to minimize $a_8$, the optimal value for $\beta_{\set{E}}$ is given by
$$
\beta_{\set{E}} = \frac{\sigma_{\set{K}}^2}{\sigma_{\set{E}}^2}.
$$
In other words, incorporating edge vectors is helpful for reducing the velocity of the third dimension of the rotational component. 

Similar analysis can be done for other $a_i$. As the rationale is similar, we omit them for brevity. 

\noindent\textbf{Contributions of keypoints, edge vectors, and symmetry correspondences.} It is very interesting to study the structure of (\ref{Eq:LSA:10}). First of all, all elements are relevant to keypoints. Edge vectors provide full constraints on the underlying rotation. Symmetry correspondences also provide constraints on two dimensions of the underlying rotation. However,  by analyzing the structure of $H_{\set{E}}$ and $H_{\set{K}}$, one can see that they do not provide constraints on two dimensions of the underlying translation (albeit on this simple model). This explains why only using edge vectors and symmetry correspondences leads to poor results on object translations.

\subsection{Proof of Propositions}
\label{Section:Proof:Propositions}

\subsubsection{Proof of Proposition~\ref{Prop:Jacobi:1}}
\label{Proof:Jacobi:1}
\noindent\textbf{Derivatives of $\bs{r}_k^{\set{K}}$ and $\bs{r}_e^{\set{E}}$.}
It is straightforward to compute the derivatives of $\bs{r}_k^{\set{K}}$ and $\bs{r}_e^{\set{E}}$ with respect to $\bs{y}_k$ and $\bs{y}_e$, respectively. In the following, we focus on the derivatives of $\bs{r}_k^{\set{K}}$ with respect to $(\bs{c},\overline{\bs{c}})$. The derivatives of $\bs{r}_e^{\set{E}}$ can be obtained by subtracting those of $\bs{r}_{e_s}^{\set{K}}$ and those of $\bs{r}_{e_t}^{\set{K}}$. 

Recall the local parameterization $R = \exp(\bs{c}\times)R^{\gt}$ and $\bs{t} = \bs{t}^{\gt} + \overline{\bs{c}}$. We have
\begin{align*}
\frac{\partial \overline{\bs{p}}_k}{\partial (\bs{c},\overline{\bs{c}})} & =   \frac{\partial ( \overline{\bs{p}}_k^{\gt} + \bs{c}\times  \overline{\bs{p}}_k^{\gt} + \overline{\bs{c}})}{\partial (\bs{c},\overline{\bs{c}})} \\
& = (- \overline{\bs{p}}_k^{\gt}\times, I_3).
\end{align*}
Using chain rule, we have
\begin{align*}
& \frac{\partial \bs{r}_k^{\set{K}}}{\partial (\bs{c},\overline{\bs{c}})}  = -\frac{1}{p_{k,z}^{\gt,3D}}(\left(
\begin{array}{c}
\frac{\partial p_{k,x}^{3D}}{\partial (\bs{c},\overline{\bs{c}})}\\
\frac{\partial p_{k,y}^{3D}}{\partial (\bs{c},\overline{\bs{c}})}
\end{array}
\right) -\left(
\begin{array}{c}
p_{k,x}\\
p_{k,y}
\end{array}
\right)\cdot \frac{\partial p_{k,z}^{3D}}{\partial (\bs{c},\overline{\bs{c}})}) \\
& = -\left(
\begin{array}{cccccc}
0 & 1 & -p_{k,y}^{\gt} & \frac{1}{p_{k,z}^{\gt,3D}} & 0 & 0\\
-1 & 0 & p_{k,x}^{\gt} & 0 & \frac{1}{p_{k,z}^{\gt,3D}} & 0 
\end{array}
\right) \\
&+ \left(
\begin{array}{c}
p_{k,x}^{\gt} \\
p_{k,y}^{\gt}
\end{array}
\right)\cdot \left(
\begin{array}{cccccc}
p_{k.y}^{\gt} & -p_{k,x}^{\gt} & 0 & 0 & 0 & \frac{1}{p_{k,z}^{\gt,3D}}
\end{array}
\right)
\end{align*}

\noindent\textbf{Derivatives of $r_s$.} Again using chain rule, 
we have
\begin{align*}
\frac{\partial r_s^{S}}{\partial \bs{y}_s} & = \left(
\begin{array}{c}
(\hat{\bs{q}}_{s1}^{\gt}\times \bs{e}_1)^T \bs{n}^{\gt} \\
(\hat{\bs{q}}_{s1}^{\gt}\times \bs{e}_2)^T \bs{n}^{\gt}
\end{array}
\right) = \left(
\begin{array}{c}
n_y^{\gt} - n_z^{\gt}q_{s1,y}^{\gt} \\
-(n_x^{\gt} - n_z^{\gt} q_{s1,x}^{\gt})
\end{array}
\right)
\end{align*}
Moreover,
\begin{align*}
\frac{\partial r_s^{S}}{\partial \bs{c}} & = \frac{\partial \det ((\hat{\bs{q}}_{s1}^{\gt}\times\hat{\bs{q}}_{s2}^{\gt},\bs{c},\bs{n} ))}{\partial \bs{c}} = \bs{n}\times (\hat{\bs{q}}_{s1}^{\gt}\times \hat{\bs{q}}_{s2}^{\gt}).
\end{align*}

\subsubsection{Proof of Proposition~\ref{Prop:GIMF}}
\label{Proof:Prop:GIMF}

\noindent\textsl{Proof:} First of all, any optimal solution $\bs{x}^{\star}(\bs{y})$ is a critical point of $f$. Therefore, it shall satisfy:
\begin{equation}
\frac{\partial f}{\partial \bs{x}} (\bs{x}^{\star}, \bs{y}) = 0.
\label{Eq:LSA:3}
\end{equation}
Consider a neighborhood, where $\|\bs{y}\|\leq \epsilon_1$, and $\|\bs{x}\|\leq \epsilon_2$. $\epsilon_2$ is chosen so that it contains for each $\bs{y}$, the critical point with the smallest norm. Assume that $\frac{\partial^2 f}{\partial^2 \bs{x}}$ is positive semidefinite in this neighborhood. 

By contradiction, suppose there exists two distinctive local minimums $\bs{x}_1(\bs{y})$ and $\bs{x}_1(\bs{y})$ for a given $\bs{y}$, i.e.,
\begin{equation}
\frac{\partial f}{\partial \bs{x}} (\bs{x}_1(\bs{y}), \bs{y}) = \frac{\partial f}{\partial \bs{x}} (\bs{x}_2(\bs{y}), \bs{y}) = 0.
\label{Eq:LSA:4}
\end{equation}
Through integration, (\ref{Eq:LSA:4}) yields
\begin{align*}
0 & = \int\limits_{0}^1 \frac{\partial^2 f}{\partial^2 \bs{x}}(\bs{x}_1 + t(\bs{x}_2-\bs{x}_1),\bs{y})(\bs{x}_2-\bs{x}_1) dt \\
& = \big(\int\limits_{0}^1 \frac{\partial^2 f}{\partial^2 \bs{x}}(\bs{x}_1 + t(\bs{x}_2-\bs{x}_1),\bs{y})dt\big)\cdot (\bs{x}_2-\bs{x}_1)
\end{align*}
Since the weighted sum of positive definite matrices is also positive definite. It follows that 
$$
\int\limits_{0}^1 \frac{\partial^2 f}{\partial^2 \bs{x}}(\bs{x}_1 + t(\bs{x}_2-\bs{x}_1),\bs{y})dt\succ 0.
$$
In other words, it cannot have a zero eigenvalue, with non-zero eigenvector $\bs{x}_2-\bs{x}_1$. In other words, the critical point is unique. Since the second order derivatives are positive definite, then each critical point is also a local minimum.

Computing the derivatives of (\ref{Eq:LSA:3}) with respect to $\bs{y}$, we obtain
\begin{equation}
\frac{\partial^2 f}{\partial^2 \bs{x}}\cdot \frac{\partial \bs{x}^{\star}}{\partial \bs{y}} + \frac{\partial^2 f}{\partial \bs{x} \partial \bs{y}} = 0
\label{Eq:2}
\end{equation}

\subsubsection{Proof of Proposition~\ref{Prop:GNLS}}
\label{Proof:Prop:GNLS}

The proof is straight-forward as $\bs{r}_i(\bs{0},\bs{0}) = 0$, and
\begin{align*}
\frac{\partial (\frac{\beta_{i,1}\bs{r}_i}{\sqrt{\beta_{i,2}^2 + \|\bs{r}_i\|^2}})}{\partial \bs{x}} &= \frac{\beta_{i,1}}{\beta_{i,2}}\cdot \frac{\partial \bs{r}_i}{\partial \bs{x}}, \\
\frac{\partial (\frac{\beta_{i,1}\bs{r}_i}{\sqrt{\beta_{i,2}^2 + \|\bs{r}_i\|^2}})}{\partial \bs{y}} &= \frac{\beta_{i,1}}{\beta_{i,2}}\cdot \frac{\partial \bs{r}_i}{\partial \bs{y}}.
\end{align*}
{\small
\bibliographystyle{ieee_fullname}
\bibliography{egbib}
}